\documentclass[letterpaper, 10 pt, conference]{ieeeconf}  

\IEEEoverridecommandlockouts                               
                                                         
\usepackage{microtype}
\usepackage{booktabs}

\usepackage{hyperref}

\usepackage{amsthm}
\usepackage{amsmath}
\usepackage{amssymb}
\usepackage{mathtools}

\usepackage[utf8]{inputenc}
\usepackage[T1]{fontenc}
\usepackage{graphicx}
\usepackage{caption}
\usepackage{subcaption}
\usepackage{float}

\usepackage[capitalize,noabbrev]{cleveref}

\theoremstyle{plain}

\theoremstyle{definition}

\theoremstyle{remark}

\usepackage{alphalph}

\title{\LARGE \bf
Genetic Algorithm with Innovative Chromosome Patterns\\
in the Breeding Process
}

\author{ \parbox{3 in}{\centering Qingchuan Lyu*
         \thanks{* Department of Computer Science, Georgia Institute of Technology, Atlanta, GA, the United States \& Visa Inc., Foster City, CA, the United States}\\
         {\tt\small qlyu9@gatech.edu}}
        \hspace*{ 0.5 in}
}

\begin{document}
\maketitle
\thispagestyle{empty}
\pagestyle{empty}

\begin{abstract}
This paper proposes Genetic Algorithm with Border Trades (GAB), a novel modification of the standard genetic algorithm that enhances exploration by incorporating new chromosome patterns in the breeding process. This approach significantly mitigates premature convergence and improves search diversity. Empirically, GAB achieves up to 8× higher fitness and 10× faster convergence on complex job scheduling problems compared to standard Genetic Algorithms, reaching average fitness scores of 888 versus 106 in under 20 seconds. On the classic Flip-Flop problem, GAB consistently finds optimal or near-optimal solutions in fewer generations, even as input sizes scale to thousands of bits. These results highlight GAB as a highly effective and computationally efficient alternative for solving large-scale combinatorial optimization problems.

\end{abstract}

\section{Introduction}
The Canonical Genetic Algorithm (GA) must balance exploration and exploitation during reproduction to avoid premature convergence near the optimum while accelerating convergence. To address these challenges, Whitley (Whitley, 1989) introduced the Genitor-style algorithm, termed “steady-state” genetic algorithms by Syswerda (Syswerda, 1989). By evolving the population incrementally, steady-state GAs better retain diversity, as fewer individuals are replaced at a time. However, steady-state GAs exhibit higher variance than canonical GAs in hyperplane sampling behavior (Syswerda, 1991).

Two major issues arise with steady-state GAs: (i) the emphasis on steady-state replacement and rank-based selection prioritizes refining \textit{existing solutions} over introducing new ones. High selection pressure can still lead to premature convergence. (ii) These algorithms require careful tuning of mutation rates and replacement strategies (Whitley, 1994).

This paper introduces a novel approach to address premature convergence and improve convergence behavior: incorporating \textit{new chromosome patterns} through border trade activities to significantly enhance exploration. The Algorithm section explores the border trade concept in two problems: Flip-Flop and Job Scheduling with breaks.
\section{Algorithm}
This paper investigates the effects of border trades in Genetic Algorithms along two dimensions. First, the Flip-Flop problems illustrate both the advantages and challenges of border trades in a simple problem space, particularly as problem sizes grow. Second, the Job Scheduling problems highlight the impact of various border trade activities in a more complex problem space. As discussed in this section and further analyzed in the following sections, a seemingly weaker parent with lower fitness than the original can still produce a superior offspring due to the interplay of crossover, mutation, and randomness.
\subsection{Border Trades in Flip-Flop}
The Flip-Flop problem involves a binary string where optimal fitness is achieved when the string alternates between 0s and 1s. In Genetic Algorithms (GAs), this problem can lead to situations where two parent chromosomes are close to the ideal solution but are too similar. As a result, crossover and mutation operators often produce identical offspring, causing stagnation in suboptimal solutions or slow fitness improvement. In Figure 1, double boundaries indicate the bits selected by a random crossover point, marked by a dotted line. In this example, a one-point crossover occurs at the second bit, resulting in no change to the fitness value after the operation. Similarly, when the crossover occurs at the 1st, 3rd, or 4th bit, the reproduced child duplicates one of the parents. Across all four possible crossover points, the fitness value of the child remains identical to that of one of its parents.

\begin{figure}[ht]
\vskip 0.2in
\begin{center}
\centerline{\includegraphics[width=\columnwidth]{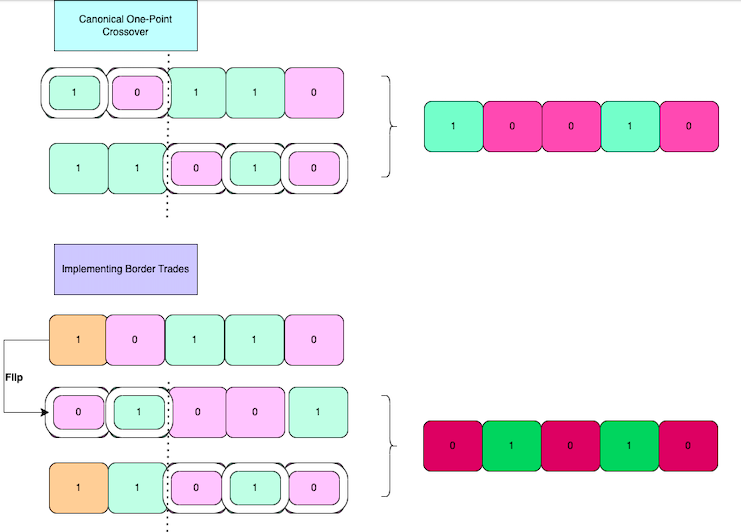}}
\caption{Border Trades with One-Point Crossover Operator}
\end{center}
\vskip -0.2in
\end{figure}

To address this, border trades (GAB) introduce new patterns to diversify the population by flipping all bits of one parent (0 becomes 1, and 1 becomes 0) when two locally optimized parent chromosomes share the same initial elements, enabling the algorithm to produce a child with a distinct bit pattern. This approach not only preserves the optimality of the parent chromosomes but also boosts diversity and exploration, allowing the population to escape suboptimal valleys.

As shown in Figure 1, implementing border trades in this scenario can produce a new chromosome with the optimal fitness value of 4. This solution is achievable when crossover occurs at the 1st or 3rd bit as well. The only failure case arises when crossover happens at the 4th bit, reducing the fitness from 3 to 2. However, this failure introduces a novel chromosome pattern with three consecutive 1's in the end, enhancing diversity and potentially leading to new solutions in subsequent generations. Overall, border trades offer a 75\% probability of producing an optimal solution in the next generation and a 25\% chance of a temporary decrease in fitness. This mechanism is conceptually similar to how hallucinations aid convergence in Q-Learning.

\subsection{Border Trades in Job Scheduling}
The Job Scheduling with Breaks problem presents a more complex scenario compared to Flip-Flop problems. The job scheduling problem dates back to the 1960s, when Duffin and Lawler proposed formulations involving profits and deadlines (Duffin \& Lawler, 1963). This paper builds upon their foundational work by introducing additional constraints, such as task duration, work limits, and mandated breaks. In this problem, an array of tasks is provided, where each task is defined by its unique ID, duration, deadline, and profit. The objective is to maximize overall profit by completing as many tasks as possible before their respective deadlines, assuming there is an unlimited demand for each task. Workers are required to take breaks: for every two units of work completed, one unit of break is mandated. For tasks with durations longer than two units, a break is taken immediately upon completing the task.

The reason why GA solutions are investigated in this problem space is brute force becomes computationally infeasible as the problem size increases. A brute-force method could optimally schedule three tasks within only 0.0017 seconds,  while the canonical Genetic Algorithm (GA) required an average of 1.75 seconds to reach the same result. But, when the problem size is 108, the brute-force method would require $108^{108}$ iterations. At a rate of 0.0001 second per iteration, it would theoretically take approximately $4.7$e+210 days to complete. Additionally, even after 6.75 million iterations, brute force achieved a fitness value of 106 in 165 seconds, whereas GA produced an average fitness value of 888 (over ten runs) in just 18 seconds. This efficiency showcases the strength of GA in complex problem spaces. 

One inherent weakness of canonical GA is the loss of task orders due to mutation and crossover operations. This is particularly detrimental in Job Scheduling problems, where deadlines are critical. Border trades mitigate this challenge by analyzing task boundaries and break periods, facilitating the exchange of tasks as required. Five border trade strategies were explored, grouped into two categories: \textit{value-based} and \textit{schedule-based} approaches.

\subsubsection{Value-Based Border Trades}
This category of strategies involve a predefined value function as follows.

\textbf{GAB-A1}: Profit, Deadline, and Duration-Based Grouping. For each task, compute the value as:
$$value = \frac{profit \times deadline}{duration}$$
Tasks are divided into two groups based on whether their value is above or below the average. To enhance diversity, every two tasks are exchanged regardless of deadlines, duration, or profit if the first task of the child chromosome belongs to the same value group as the first task of the last chromosome (similar to border trades in Flip-Flop). However, this strategy does not directly consider deadlines or overall profit, limiting its effectiveness in certain cases.

\textbf{GAB-A2}: Profit-Only Grouping. This simplified version computes the value as the profit of each task, splitting tasks into two groups based on their profit relative to the average. While less computationally intensive than GAB-A1, GAB-A2 focuses solely on profit, which can overlook other factors like deadlines and duration.

\subsubsection{Schedule-Based Border Trades}
Schedules are defined as ordered arrays containing both tasks and breaks, derived from chromosomes. Unlike chromosomes, which only include tasks, schedules account for deadlines, durations, and breaks. A single chromosome can correspond to multiple valid schedules. Three strategies were examined:

\textbf{GAB-B}: Unconditional Exchanges. Border trades define the boundary between tasks and breaks, exchanging tasks before and after each break without immediately calculating the fitness of the new schedule. The primary goal of this step is to enhance diversity, enabling the algorithm to escape local optima. As presented in Section III.B.1, this general border trade strategy achieved higher fitness scores than selective border trade strategies.

\textbf{GAB-C1}: Conditional Exchanges. Perform border trades as in GAB-B, but retain the new schedule only if it improves local fitness.

\textbf{GAB-C2}: Immediate Evaluation. Swap tasks around each break, evaluating fitness after each trade. Retain the new schedule if fitness improves; otherwise, revert the trade and proceed to the next break.

\begin{figure}[ht]
\vskip 0.2in
\begin{center}
\centerline{\includegraphics[width=\columnwidth]{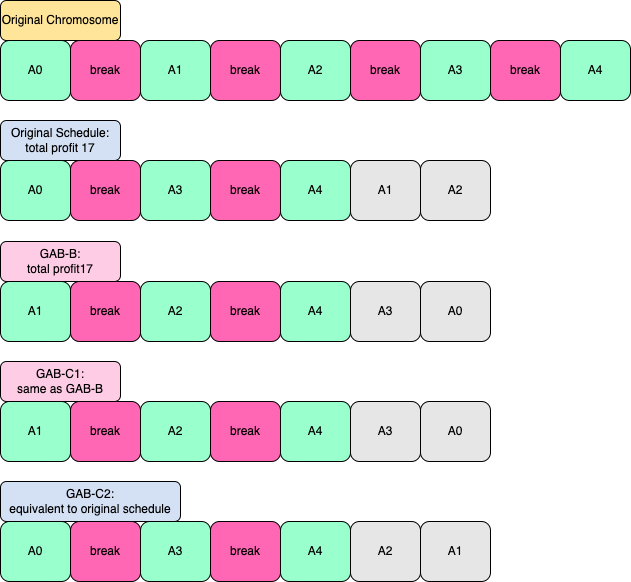}}
\caption{Illustrations of Schedule-Based Border Trade Strategies}
\end{center}
\vskip -0.2in
\end{figure}

To illustrate the differences in Figure 2, consider a set of five tasks $A_0, A_1, \dots , A_4$, where each task is assigned a tuple representing its duration, deadline, and profit as follows: $A_0: (5, 6, 8)$, $A_1: (4, 7, 7)$, $A_2: (3, 8, 6)$, $A_3: (2, 9, 5)$, $A_4: (1, 10, 4)$. Suppose one of the parent chromosomes selected for reproduction is ordered as $(A_0, A_1, A_2, A_3, A_4)$. When constructing a programmatic schedule from this ordered chromosome, the algorithm iterates sequentially through the tasks, moving those that fail to meet their respective deadline constraints to the end of the schedule (denoted by light gray in Figure 2). The corresponding schedules, incorporating border trades, are illustrated in Figure 2.

The outcomes reveal distinct behaviors across the border trade strategies. Specifically, Figure 2 demonstrates that GAB-C1 produced an identical result to GAB-B. In contrast, GAB-C2 generated a schedule equivalent to the original by rejecting the trade near the first break but successfully executing the trade at the second break. This nuanced approach highlights the conditional nature of GAB-C2, which evaluates the fitness impact of each trade before deciding whether to retain or abandon the modification.

\section{Empirical Evidence}
The analysis presented in this paper includes the evaluation of average fitness values, function evaluations (FEvals), and wall-clock time over 10 runs, with standard deviation indicated by shaded regions as part of the sensitivity analysis curves. Two distinct notions of convergence are employed throughout the paper:

\textbf{Notion of Convergence}: Convergence is defined as the point at which an algorithm reaches the maximum fitness value, with subsequent differences not exceeding 0.5 units below the maximum.

\textbf{Notion of Semi-Convergence}: In the context of the Flip-Flop and Job Scheduling problems, not all algorithms reach the highest possible fitness value. Semi-convergence is defined as the point at which the average fitness value of an algorithm, obtained from 10 runs, is at most one unit less than the desired maximum fitness value, with subsequent differences never exceeding 1 unit from the maximum.

In this study, the term ``does not converge" (denoted as ``N/A" in tables) refers to the situation where algorithms fail to converge within 2048 iterations or fail to improve fitness values within 500 maximum attempts. When an algorithm reaches 500 unsuccessful attempts to improve fitness values or exceeds the iteration limit of 2048, it is terminated.

\subsection{Analysis of Flip-Flop Problem Results}
Population sizes and mutation rates were jointly tuned during the evaluation of GA and GAB across problem sizes: 1000, 28, 14, and 7. Detailed hyperparameter tuning processes and results are in Appendix A.1. The primary objective was to strike a balance — selecting population sizes small enough to capture the subtle nuances of alternation windows while ensuring they were sufficiently large to prevent excessive computational overhead and runtime. 

In the Flip-Flop problem, the performance of GA and GAB varied significantly across different problem sizes. Specifically, GA failed to achieve semi-convergence, while GAB successfully converged at a problem size of 1,000. At a size of 28, GA achieved semi-convergence, whereas GAB attained full convergence. Notably, at a size of 7, both GA and GAB converged; however, across all four problem sizes, GAB consistently achieved full convergence in fewer iterations compared to both GA and Simulated Annealing (SA), demonstrating its superior efficiency in convergence performance, as Table I summarized. For clarity, the term "size" refers to the length of the binary string being evaluated.

\begin{table}[h]
\caption{Comparing Convergence Iterations Among SA, GA and GAB on Flip-Flop. }
\vskip 0.15in
\begin{center}
\begin{small}
\begin{sc}
\begin{tabular}{lcccr}
\toprule
Sizes & 7 & 14 & 28 & 1,000 \\
\midrule
SA & 51 & 1208 & 1258 & N/A\\
GA & 89 & N/A & N/A & N/A\\
\textbf{GAB} & \textbf{36} & \textbf{51} & \textbf{92} & \textbf{981}\\
\bottomrule
\end{tabular}
\end{sc}
\end{small}
\end{center}
\vskip -0.1in
\end{table}

The subsequent sections provide a detailed comparison of the performance of GA and GAB across different problem sizes. A comprehensive picture of SA's fitness curves can be found in Appendix C.

\subsubsection{Analysis at Problem Size 1,000}
\begin{figure}[ht]
\vskip 0.2in
\begin{center}
\begin{subfigure}[b]{0.22\textwidth}
         \centering
         \includegraphics[width=\linewidth]{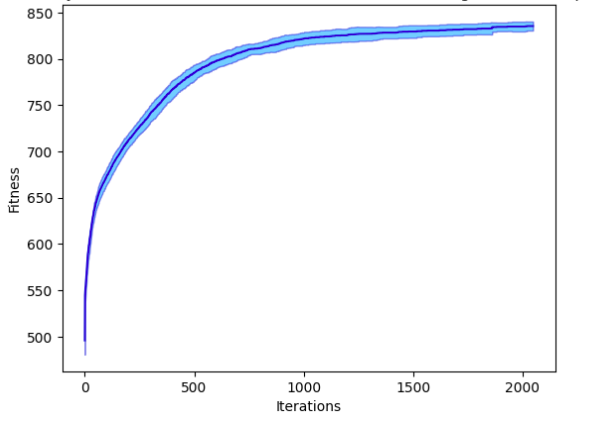}
         \caption{GA Fitness by Iterations}         
     \end{subfigure}
     \hfill
     \begin{subfigure}[b]{0.22\textwidth}
         \centering
         \includegraphics[width=\linewidth]{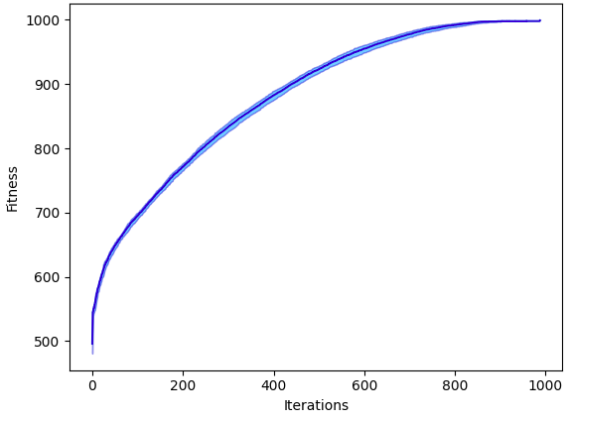}
         \caption{GAB Fitness by Iterations}
     \end{subfigure}
     \hfill
     \begin{subfigure}[b]{0.22\textwidth}
         \centering
         \includegraphics[width=\linewidth]{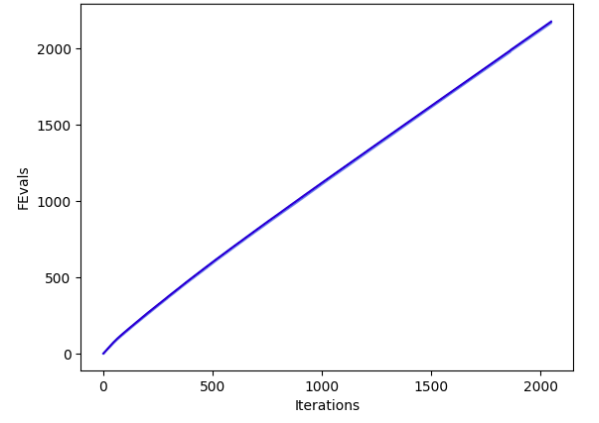}
         \caption{GA FEvals by Iterations}         
     \end{subfigure}
     \hfill
     \begin{subfigure}[b]{0.22\textwidth}
         \centering
         \includegraphics[width=\linewidth]{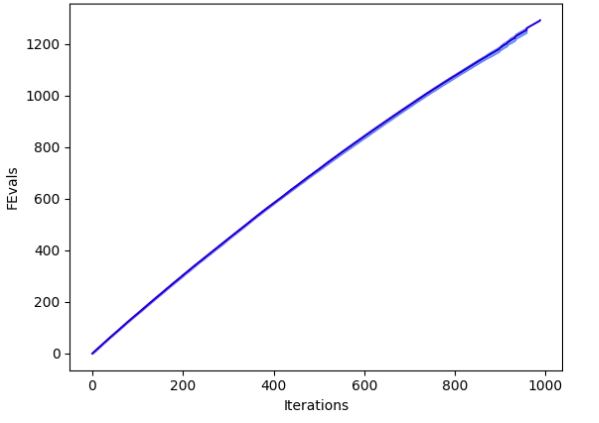}
         \caption{GAB FEvals by Iterations}
     \end{subfigure}
     \hfill
     \begin{subfigure}[b]{0.22\textwidth}
         \centering
         \includegraphics[width=\linewidth]{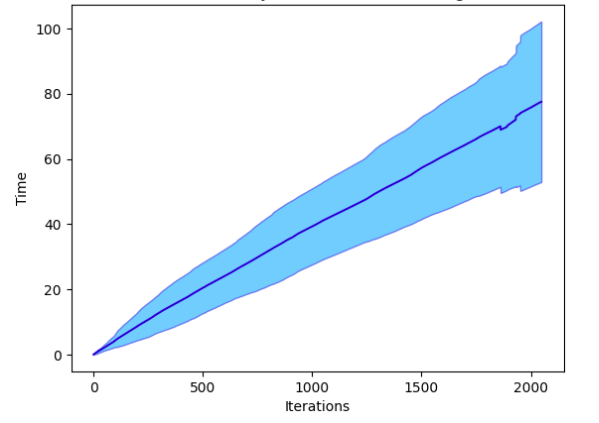}
         \caption{GA Time by Iterations}         
     \end{subfigure}
     \hfill
     \begin{subfigure}[b]{0.22\textwidth}
         \centering
         \includegraphics[width=\linewidth]{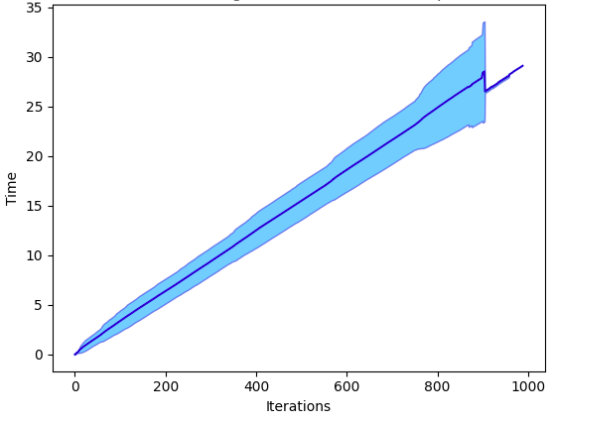}
         \caption{GAB Time by Iterations}
     \end{subfigure}
     \hfill
     \caption{Genetic Algorithm (GA) versus Genetic Algorithm with Border trades (GAB): Performance comparison in Flip-Flop problem of size 1,000}   
\vskip -0.2in
\end{center}
\end{figure}

In the problem space with size = 1,000, only GAB achieved the optimal fitness value of 999 within 2,048 iterations. In contrast, GA failed to converge, becoming trapped in suboptimal solutions, with its highest recorded average fitness value reaching only 835.4, whereas the highest individual fitness achieved by GA was 843. The fitness progression for GA followed a logarithmic trend, exhibiting slow increments beyond the 500th iteration. Figure 3(a) showed GA achieved a fitness increase of 289 during the first 500 iterations but slowed dramatically, gaining only 50 additional fitness points over the subsequent 1,548 iterations. Conversely, GAB displayed a consistently steeper slope in its fitness curve, demonstrating a significantly higher rate of improvement, as shown in Figure 3(b). This outcome suggests that, similar to its performance in the problem space with size = 28, the canonical GA lacked the sophistication necessary for larger problem spaces. 

The standard deviation (Std Dev) of GA's fitness curve ranged from 5 to 15 throughout the iterations. The largest Std Dev occurred at iteration 0 due to random initiations, gradually declining over time, with Std Dev 9 at iteration 100, Std Dev 6 at iteration 800, and Std Dev 5 at iteration 2,000. This indicates that GA lacked the exploratory power necessary to escape suboptimal valleys and approach the desired fitness value of 999. These results underscore that the canonical GA is not robust enough for larger problem spaces and demonstrates increasingly poor convergence behavior as the problem space grows.

In terms of function evaluations (FEvals), both GA and GAB exhibited similar linear upward trends and minimal standard deviations (Std Dev) in the problem space of size = 1,000, consistent with observations for size = 28. However, during 95\% of the 989 valid iterations of GAB, it required slightly more FEvals than GA. By the final iteration (989th), GAB averaged 1,292 FEvals compared to GA’s 1,106. In the 43 iterations where GA required equal to or slightly more function evaluations than GAB, the difference was negligible, never exceeding 1.11. On average, GAB performed 103 more FEvals per iteration than GA. This observation reveals that GA required fewer computational efforts overall by design.

Notably, GA expended 2,174 FEvals on average by the 2,048th iteration without approaching the optimal fitness value, whereas GAB reached convergence at the desired optimal fitness with only 1,255 FEvals by the 989th iteration. These results demonstrate that GAB was significantly more efficient than GA in terms of both convergence speed and achieving optimal fitness values.

In terms of wall-clock runtime, GA required an average of 78 seconds to complete 2,048 iterations, failing to achieve fitness values close to optimal. In contrast, GAB required only 29 seconds to converge to the optimal solution in 989 iterations. Furthermore, GAB consistently completed each of the first 989 iterations faster than GA, with GA averaging 4.7 seconds more than GAB per iteration. This suggests that the additional computational steps involved in GAB’s border trade mechanism did not result in a time penalty. Instead, the benefits of increased exploration and diversity outweighed the time cost of the enhanced algorithmic structure.

Overall, GAB outperformed GA in both time efficiency and the number of iterations required to achieve the desired optimal fitness value, while GAB incurred a higher FEval cost per iteration due to its more sophisticated structure. By comparison, GA failed to approach the optimal solution within the iteration boundaries, highlighting its limitations in handling larger problem spaces.

\subsubsection{Analysis at Problem Size 28}
\begin{figure}[ht]
\vskip 0.2in
\begin{center}
\begin{subfigure}[b]{0.22\textwidth}
         \centering
         \includegraphics[width=\textwidth]{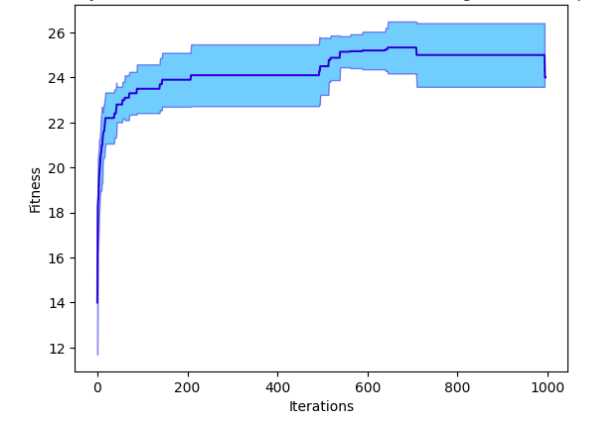}
         \caption{GA Fitness by Iterations}         
     \end{subfigure}
     \hfill
     \begin{subfigure}[b]{0.22\textwidth}
         \centering
         \includegraphics[width=\textwidth]{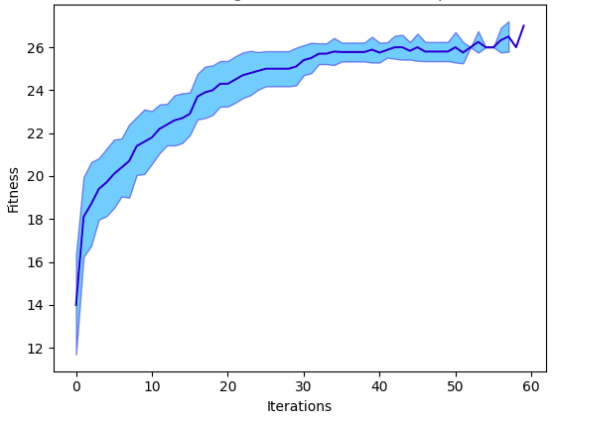}
         \caption{GAB Fitness by Iterations}
     \end{subfigure}
     \hfill
     \begin{subfigure}[b]{0.22\textwidth}
         \centering
         \includegraphics[width=\textwidth]{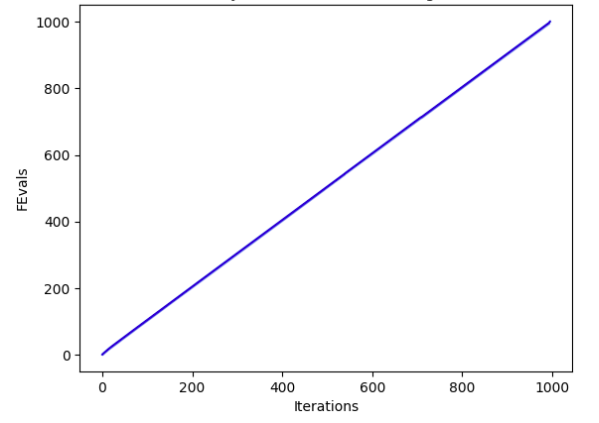}
         \caption{GA FEvals by Iterations}         
     \end{subfigure}
     \hfill
     \begin{subfigure}[b]{0.22\textwidth}
         \centering
         \includegraphics[width=\textwidth]{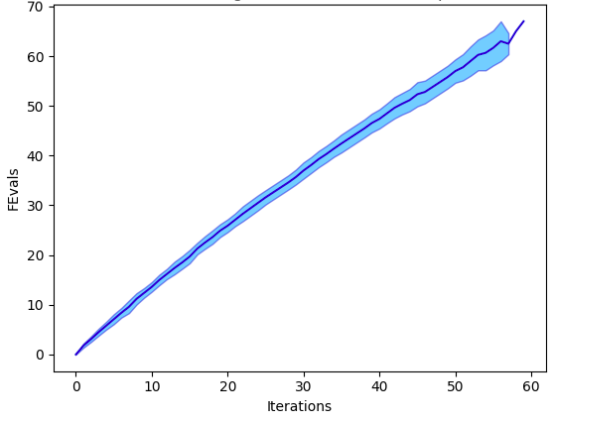}
         \caption{GAB FEvals by Iterations}
     \end{subfigure}
     \hfill
     \begin{subfigure}[b]{0.22\textwidth}
         \centering
         \includegraphics[width=\textwidth]{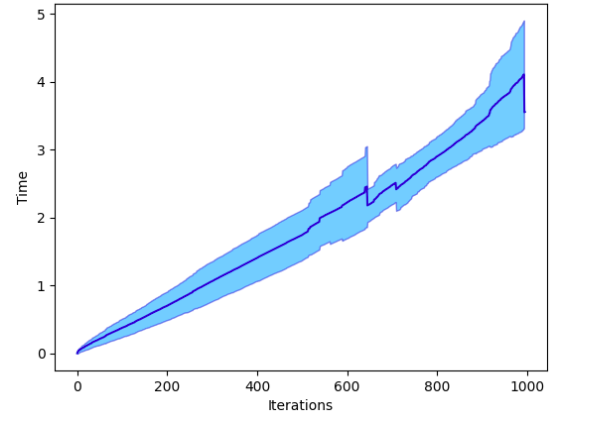}
         \caption{GA Time by Iterations}         
     \end{subfigure}
     \hfill
     \begin{subfigure}[b]{0.22\textwidth}
         \centering
         \includegraphics[width=\textwidth]{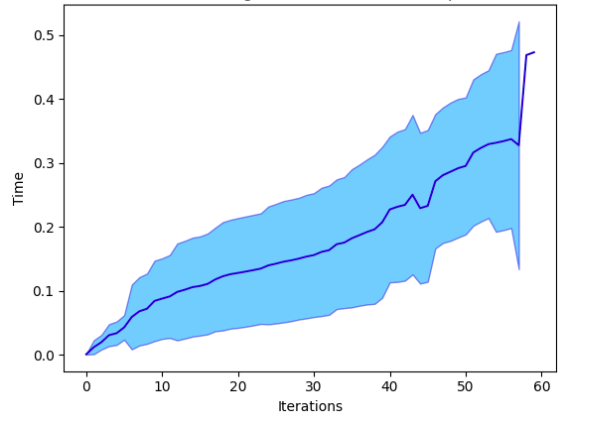}
         \caption{GAB Time by Iterations}
     \end{subfigure}
     \hfill
     \caption{Genetic Algorithm (GA) versus Genetic Algorithm with Border trades (GAB): Performance comparison in Flip-Flop problem of size 28}   
\vskip -0.2in
\end{center}
\end{figure}

For a problem size of 28, GAB demonstrated a significant advantage over GA in terms of convergence. Specifically, GAB reached the optimal fitness value of 27 at the 59th iteration, while GA only achieved semi-convergence after 539 iterations and failed to fully converge within the maximum limit of 2048 iterations. This outcome underscores the effectiveness of the border-trades mechanism in improving convergence. Furthermore, GAB achieved higher fitness values more rapidly: it achieved an average fitness value of 26 within 42 iterations over 10 runs, while GA never reached this level, with its highest average fitness value being 25.3 over 10 runs.

In terms of Function Evaluations (FEvals) per iteration, both GA and GAB exhibited similar linear trends. However, GAB displayed slightly greater variability in FEvals, with a maximum standard deviation of 4 compared to GA’s 2.12 prior to GAB's convergence iteration, as shown in Figures 4[c] and 4[d]. This variability reflects the exploratory benefit of the border-trades mechanism, which helped GAB avoid getting trapped in suboptimal solutions during the early stages. While GA required fewer FEvals than GAB in 97\% of the first 60 iterations — averaging 2.4 fewer FEvals per iteration with a maximum difference of 3.9 — this marginal efficiency was insufficient to overcome GA’s inability to converge. Notably, at the 995th iteration, where all 10 GA runs terminated due to the failure to walk out of a suboptimal valley within 500 attempts, GA was consuming 1001 FEvals on average. In contrast, GAB reached convergence at the 59th iteration with only 67 FEvals, highlighting its computational efficiency in achieving optimal solutions.

The run-time trends at problem size 28 mirrored those observed for FEvals. At the point of convergence (59th iteration), GAB required an average runtime of 0.47 seconds across 10 runs, which was more than twice the 0.2 seconds that GA required for the same number of iterations. GAB also exhibited slightly higher runtime variability, with a maximum standard deviation of 0.19 seconds compared to GA’s 0.077 seconds. This additional computational cost aligns with the extra steps introduced by the border-trades mechanism. On average, GAB took 0.099 seconds longer than GA per iteration during the first 60 iterations, with runtime differences ranging from 0.00024 seconds to 0.329 seconds.

Despite the higher computational cost per iteration, GAB achieved convergence at the optimal fitness value within 0.47 seconds, whereas GA required 1.5 seconds to only semi-converge at the suboptimal fitness value of 26. This indicates that the additional computational complexity introduced by border trades was outweighed by the improvements in exploration and convergence efficiency. 

\subsubsection{ANALYSIS AT PROBLEM SIZE 7}
Table I highlights the efficiency of border trades not only in large problem spaces but also in accelerating convergence behavior in smaller problem spaces, such as a size of 7. In this study, GA achieved convergence by the 89th iteration with an average runtime of 0.1 seconds per iteration, whereas GAB required only 36 iterations to reach convergence. Interestingly, both GA and GAB achieved semi-convergence at similar points: GA by the 10th iteration and GAB by the 6th iteration. However, it should be noted that at iterations 36, 41, and 70, GA exhibited minor fluctuations in fitness values before falling back into suboptimal valleys, as Figure 5[a] shows. This behavior suggests that GA lacks a robust mechanism for escaping suboptimal regions, whereas GAB's border-trading mechanism provided the exploratory capability needed to identify the correct direction toward the optimal solution, as Figure 5[b] shows.

Boosted by border trades, GAB exhibited proactive search behavior, reflected in an average standard deviation of 0.65 when it was not trapped in suboptimal regions - greater than the GA standard deviation of 0.54. This finding aligns with the Reinforcement Learning principle that exploration (or "hallucinations") facilitates convergence to the optimal solution.

\begin{figure}[ht]
\vskip 0.2in
\begin{center}
\begin{subfigure}[b]{0.22\textwidth}
         \centering
         \includegraphics[width=\textwidth]{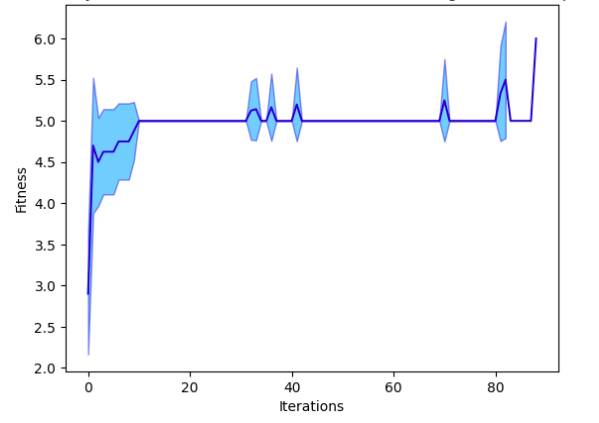}
         \caption{GA Fitness by Iterations}         
     \end{subfigure}
     \hfill
     \begin{subfigure}[b]{0.22\textwidth}
         \centering
         \includegraphics[width=\textwidth]{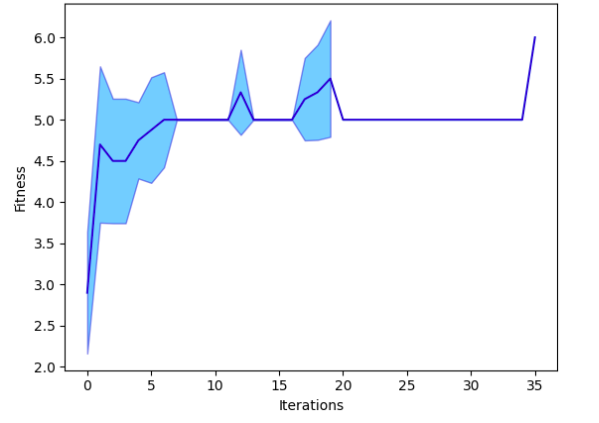}
         \caption{GAB Fitness by Iterations}
     \end{subfigure}
     \hfill
     \begin{subfigure}[b]{0.22\textwidth}
         \centering
         \includegraphics[width=\textwidth]{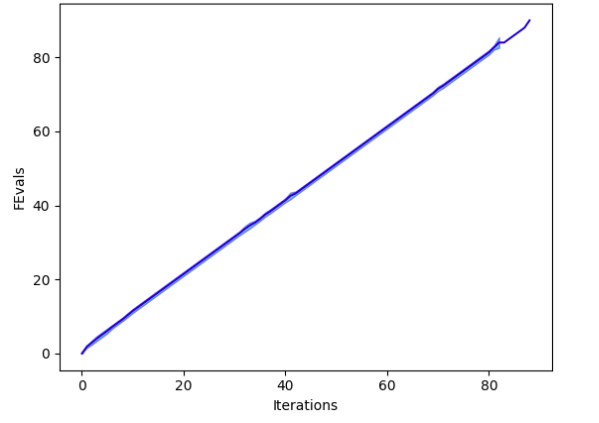}
         \caption{GA FEvals by Iterations}         
     \end{subfigure}
     \hfill
     \begin{subfigure}[b]{0.22\textwidth}
         \centering
         \includegraphics[width=\textwidth]{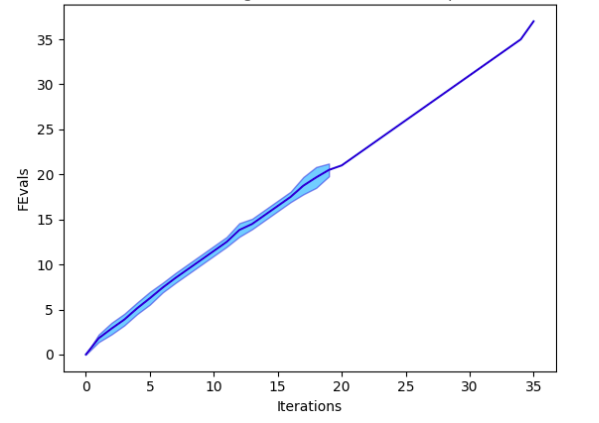}
         \caption{GAB FEvals by Iterations}
     \end{subfigure}
     \hfill
     \begin{subfigure}[b]{0.22\textwidth}
         \centering
         \includegraphics[width=\textwidth]{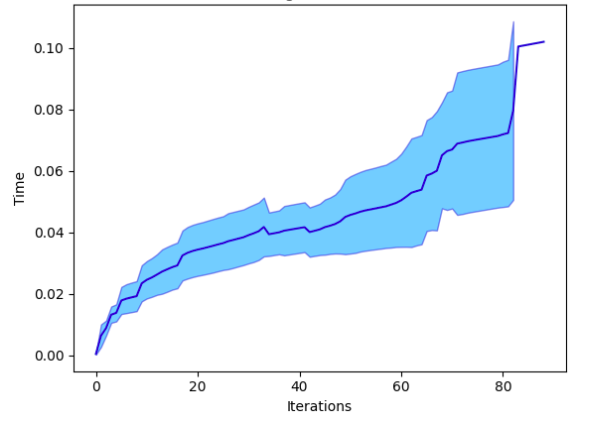}
         \caption{GA Time by Iterations}         
     \end{subfigure}
     \hfill
     \begin{subfigure}[b]{0.22\textwidth}
         \centering
         \includegraphics[width=\textwidth]{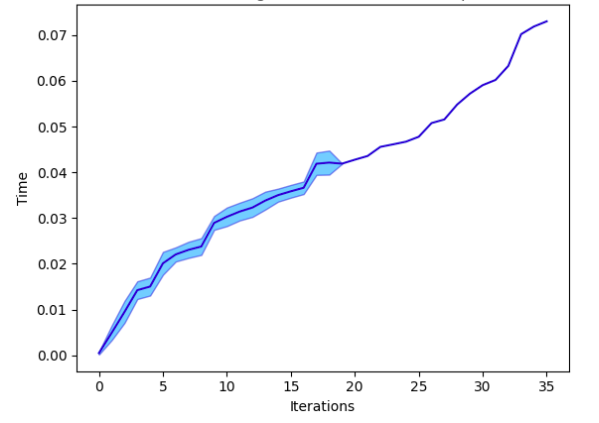}
         \caption{GAB Time by Iterations}
     \end{subfigure}
     \hfill
     \caption{Genetic Algorithm (GA) versus Genetic Algorithm with Border trades (GAB): Performance comparison in Flip-Flop problem of size 7}   
\vskip -0.2in
\end{center}
\end{figure}

Despite differences in iteration counts and runtime, the function evaluations (FEvals) for GA and GAB were remarkably similar before GAB converged. For example, at the 35th iteration, GAB required an average of 37 FEvals across 10 runs, compared to GA's 36.3 FEvals. This suggests that the additional computational cost in GAB was relatively smaller in the smaller problem space, which is logical, as there are fewer bits involved in the border-trading process.

The wall-clock run-time of GA and GAB showed distinct patterns in Figure 5[e] and 5[f]. While both algorithms converged within a similar total runtime — 0.1 seconds for GA and 0.07 seconds for GAB — there were notable differences in per-iteration efficiency: GA required 0.04 seconds on average for the 35th iteration, which is half the runtime that GAB needed. However, GAB converged in far fewer iterations, emphasizing its overall efficiency. Furthermore, the standard deviation of runtime for GAB ranged between 4.1e-05 seconds and 0.0026 seconds, which was generally smaller than GA's range of 0.0001 seconds to 0.029 seconds. This suggests that GAB maintained more consistent progress toward convergence, whereas GA experienced erratic movements that frequently reverted to suboptimal valleys, as Figure 5[a] shows.

In summary, GAB demonstrated more efficient and consistent behavior compared to GA, even in small problem spaces. Its border-trading mechanism enabled faster convergence, better exploratory performance, and consistent computational efficiency, whereas GA's lack of such mechanisms left it prone to stagnation in suboptimal regions despite comparable FEvals per iteration.

\subsection{Analysis of Job Scheduling Problem Results}
The Job Scheduling problem, characterized by more input parameters and constraints than the Flip-Flop problem, was used to evaluate the effectiveness of different border trading strategies. Subsequently, the GA that incorporates the most effective border trade strategy will be compared with the standard GA in various problem sizes within a smaller problem space.

\subsubsection{Performance Comparison of Various Border Trading Strategies}
\begin{table}[h]
\caption{Performance of GA with Various Border Trade Strategies: Each column presents the maximum average values across 10 runs.}
\vskip 0.15in
\begin{center}
\begin{small}
\begin{sc}
\begin{tabular}{lcccr}
\toprule
Algorithms & Fitness & Time (s) & FEvals \\
\midrule
GA    & 888 & \textbf{18} & \textbf{2,139}\\
GAB-A1 & 906.7 & 24.7 & 2,142.6\\
GAB-A2 & 905 & 19 & 2,146\\
\textbf{GAB-B}    & \textbf{917} & 43 & 2,146\\
GAB-C1    & 912  & 48 & 202,859\\
GAB-C2     & 784  & 33 & 214,353\\
\bottomrule
\end{tabular}
\end{sc}
\end{small}
\end{center}
\vskip -0.1in
\end{table}
To assess the potential of various border trading strategies, algorithms use the same set of 108 randomly generated tasks, representing a particularly challenging scenario where a brute-force method would need to evaluate $108^{108}$ combinations. Table II summarizes the progress of each algorithm within 2048 iterations and 500 max attempts. It demonstrates that GAB-B achieved the highest fitness values, with Function Evaluations (FEvals) being a close third place and run-time being acceptable despite the added computational cost of the border trading step. A comprehensive view of performance comparison is in Appendix B.1.

Looking at value-based strategies, both GAB-A1 and GAB-A2 outperformed the canonical GA regarding fitness, with similar FEvals. However, GAB-A2 showed slightly worse performance in both fitness and FEvals, as illustrated in Table II, although its runtime was faster due to simplified value computation. In contrast, GAB-C1 and GAB-C2 exhibited FEvals that were approximately 100 times FEvals of the other methods, indicating that these strategies often became trapped at the same suboptimal fitness value. This suggests that restricting border trades to instances where an individual chromosome achieves a better fitness score hindered sufficient exploration, resulting in GAB-C1 and GAB-C2 becoming stuck. Particularly, the lowest fitness value achieved by GAB-C2 implies that border trades at a single break point were insufficient to increase diversity or steer the algorithm toward an optimal direction, as illustrated in the example shown in Appendix B.1 Figure 7[p]. Conversely, indiscriminately swapping every two tasks, without considering break points or task durations, impeded the effectiveness of value-based strategies in meeting deadlines and maximizing profits.

\begin{table}[h]
\caption{Elbow Analysis: Iterations, Average Fitness Values, and Standard Deviation of Fitness at Turning Points}
\vskip 0.15in
\begin{center}
\begin{small}
\begin{sc}
\begin{tabular}{lcccr}
\toprule
Algorithms & Iteration & Fitness & Std Dev \\
\midrule
GA & 31 & 536.5 & 39 \\
GAB-A1 & 85 & 647.8 & 41 \\
GAB-A2 & 81 & 613.2 & 34 \\
GAB-B & 141 & 725 & \textbf{81.8} \\
GAB-C1 & \textbf{147} & \textbf{744} & 50 \\
GAB-C2 & 32 & 460 & 24 \\
\bottomrule
\end{tabular}
\end{sc}
\end{small}
\end{center}
\vskip -0.1in
\end{table}

While all fitness curves exhibited a logarithmic pattern in Appendix B.1, their long tails began to flatten at different iterations. Modifications to the value functions did not result in significant differences in performance. GAB-A1 and GAB-A2 displayed similar turning points in Table III, with GAB-A1 reaching a flatter tail in a later iteration at a higher fitness value and a wider standard deviation. Given that GAB-A1 also achieved a slightly higher overall fitness value, this suggests that incorporating task deadlines and durations into the value function definitions in GAB-A1 had a modest effect on improving search behavior.

In contrast, when considering schedules, GAB-B and GAB-C1 exhibited similar fitness curve shapes, with elbow points at 141 and 147 iterations, respectively. The average fitness values at these points were 725 and 744 for the two strategies. GAB-B achieved a slightly higher average fitness overall (917 vs. 912), which can be attributed to the additional exploration facilitated by hallucinations. In contrast, GAB-C1 experienced a reduction in exploratory behavior due to the use of a new schedule with higher fitness values only, resulting in significantly higher (approximately 100 times) function evaluations compared to GAB-B.

Conversely, the canonical GA and GAB-C2 demonstrated the weakest performance in Table III. Both strategies shared similar turning points at 31 and 32 iterations, but GAB-C2 flattened out at a lower average fitness (460 vs. 536.5) and a smaller standard deviation of 24, compared to 39 for GA. This highlights a classic case of overtuning: while GAB-B and GAB-C1 considered border trades at all breakpoints, GAB-C2 focused on a single breakpoint at a time and reversed trades when no fitness improvement was observed. This overly ``sophisticated" mechanism led to the highest function evaluations and, ultimately, worse overall performance compared to the canonical GA. As shown in Table III, GAB-B exhibited the highest standard deviation of fitness at the elbow point, which was notably greater than other strategies, further emphasizing how exploration can increase the likelihood of reaching better fitness values in the end.

\subsubsection{Performance of GAB-B and GA}
Given that GAB-B demonstrated superior performance compared to other border trade strategies, as discussed in the previous section, this section will focus on a comparative analysis of GAB-B and GA. Table IV illustrates that GAB-B significantly outperformed GA in terms of convergence iterations, wall-clock time and function evaluations. The corresponding hyperparameter tuning details are provided in Appendix A.2. This disparity underscores the superior efficiency of GAB-B in navigating complex problem spaces. The following discussion further elaborates on these findings.

Notably, GA required 501 iterations to converge at a problem size of 3, which decreased to 72 iterations when the problem size increased to 7, before subsequently increasing as the problem size grew. In contrast, GAB-B successfully converged within 135 iterations for problem sizes up to 13, exhibiting a generally stable and slightly increasing trend, as detailed in Table IV. Given that the hyperparameter candidates for GA and GAB-B were identical at each problem size, these results suggest that the incorporation of border trades enhanced GA's stability across varying problem sizes and reduced its sensitivity to specific hyperparameters. Moreover, the unusually large convergence iteration of GA at a problem size of 3 can be partially attributed to the unusually large standard deviation of its fitness curve. As depicted in Figure 6[a], GA’s fitness curve exhibited a standard deviation greater than 10 for 188 iterations, whereas GAB-B’s fitness curve exceeded this threshold for only 8 iterations. Furthermore, the average fitness of GA stagnated at 178 for a long time between iterations 233 and 500. This suggests that GA lacked a consistent search direction in certain runs, leading to prolonged entrapment in suboptimal valleys, while in other runs, it was able to escape more quickly. This observation highlights the instability of GA’s performance across the 10 runs, as it failed to ensure sufficient exploration from the outset.

\begin{figure}[ht]
\vskip 0.2in
\begin{center}
\begin{subfigure}[b]{0.22\textwidth}
         \centering
         \includegraphics[width=\textwidth]{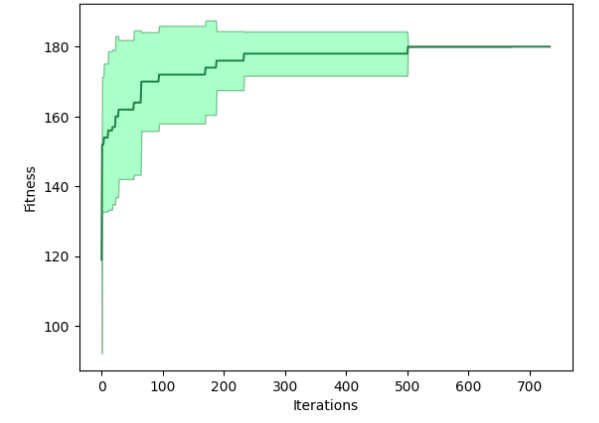}
         \caption{GA Fitness, Size=3}         
     \end{subfigure}
     \hfill
     \begin{subfigure}[b]{0.22\textwidth}
         \centering
         \includegraphics[width=\textwidth]{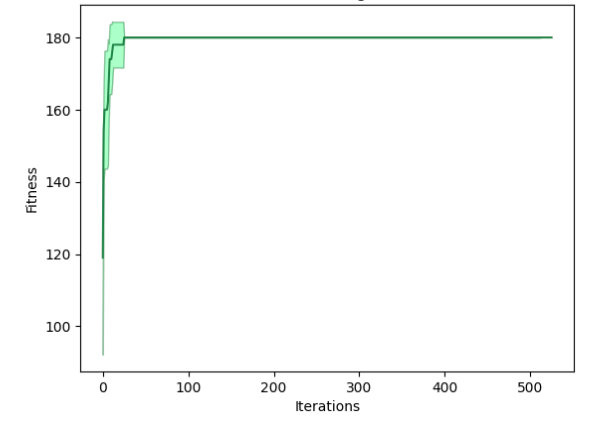}
         \caption{GAB-B Fitness, Size=3}
     \end{subfigure}
     \hfill
     \begin{subfigure}[b]{0.22\textwidth}
         \centering
         \includegraphics[width=\textwidth]{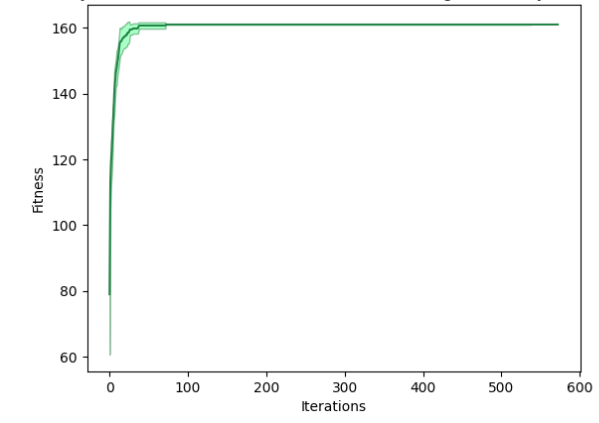}
         \caption{GA Fitness, Size=7}         
     \end{subfigure}
     \hfill
     \begin{subfigure}[b]{0.22\textwidth}
         \centering
         \includegraphics[width=\textwidth]{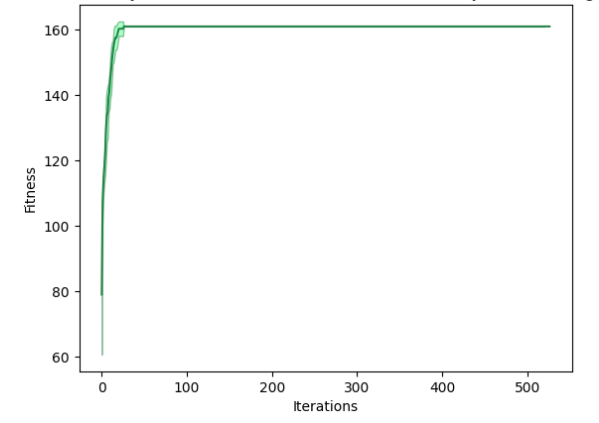}
         \caption{GAB-B Fitness, Size=7}
     \end{subfigure}
     \hfill
     \begin{subfigure}[b]{0.22\textwidth}
         \centering
         \includegraphics[width=\textwidth]{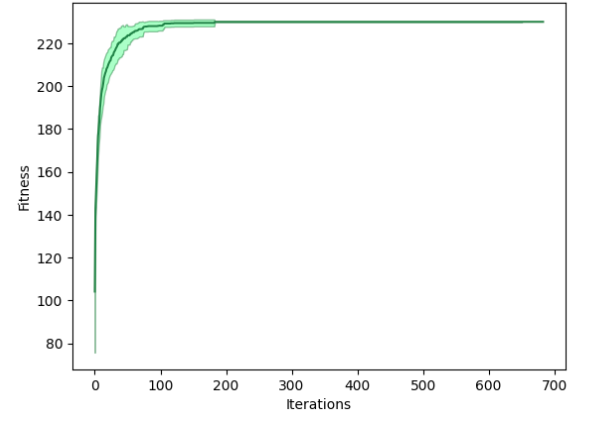}
         \caption{GA Fitness, Size=10}         
     \end{subfigure}
     \hfill
     \begin{subfigure}[b]{0.22\textwidth}
         \centering
         \includegraphics[width=\textwidth]{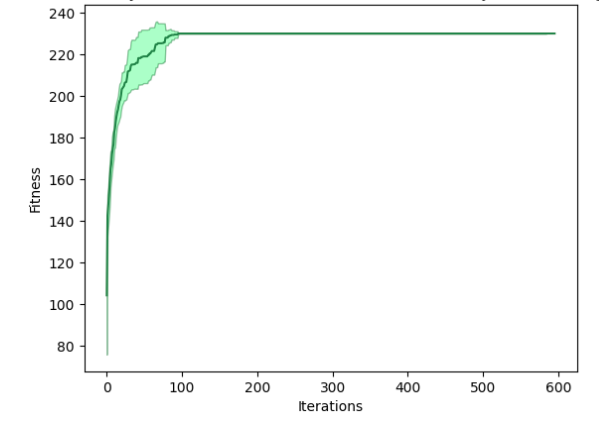}
         \caption{GAB-B Fitness, Size=10}
     \end{subfigure}
     \hfill
          \begin{subfigure}[b]{0.22\textwidth}
         \centering
         \includegraphics[width=\textwidth]{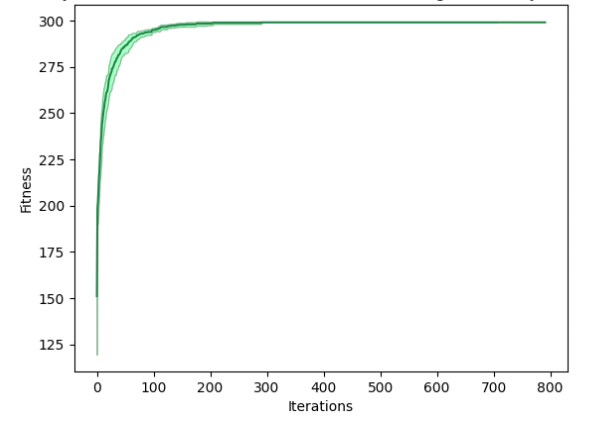}
         \caption{GA Fitness, Size=13}         
     \end{subfigure}
     \hfill
     \begin{subfigure}[b]{0.22\textwidth}
         \centering
         \includegraphics[width=\textwidth]{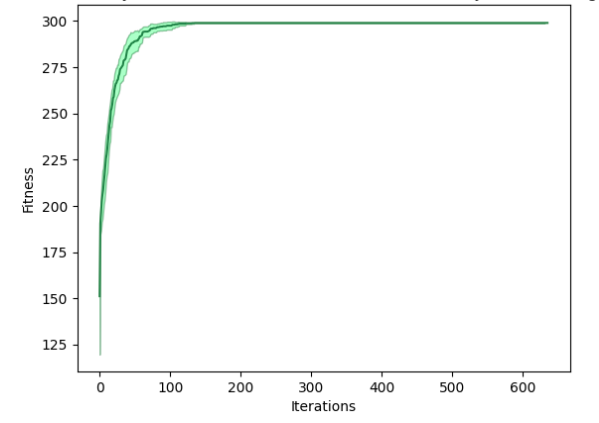}
         \caption{GAB-B Fitness, Size=13}
     \end{subfigure}
     \hfill
     \begin{subfigure}[b]{0.22\textwidth}
         \centering
         \includegraphics[width=\textwidth]{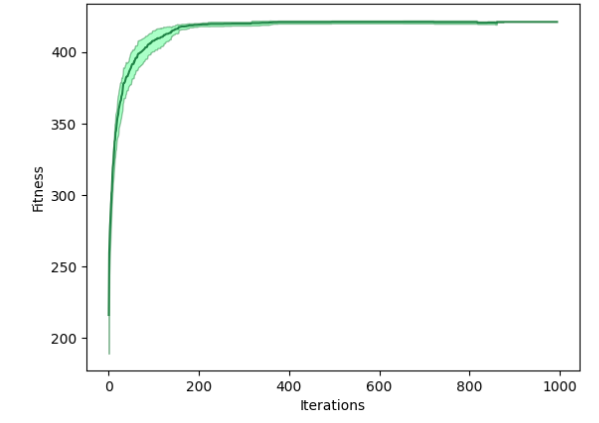}
         \caption{GA Fitness, Size=18}         
     \end{subfigure}
     \hfill
     \begin{subfigure}[b]{0.22\textwidth}
         \centering
         \includegraphics[width=\textwidth]{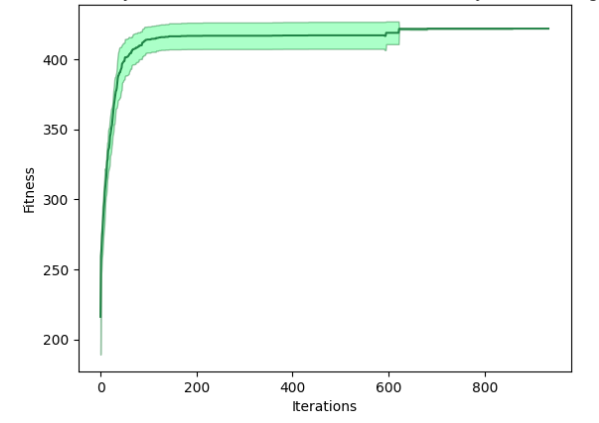}
         \caption{GAB-B Fitness, Size=18}
     \end{subfigure}
     \hfill
     \caption{Fitness Curve Comparisons Between GA and GAB-B in Job Scheduling with Breaks problems. Left: GA; right: GAB-B.}   
\end{center}
\vskip -0.2in
\end{figure}

In additional to fewer convergence iterations at each problem size in this study, GAB-B required less computational time and fewer function evaluations to achieve convergence in Job Scheduling with Breaks problems, as presented in Table IV. The “hallucination” effect induced by blind border trades at each break time enhanced GAB-B’s ability to explore the search space more effectively and attain higher fitness values. The computational overhead introduced by the border trade steps was offset by the improved search efficiency. A comprehensive graphical summary of GA and GAB-B performance metrics is provided in Appendix B.2.

\begin{table}[ht]
\caption{Comparison of the Convergence Performance Between GA and GAB-B}
\vskip 0.15in
\begin{center}
\begin{small}
\begin{sc}
\begin{tabular}{lcccr}
\toprule
Metrics: & Size & Iteration & Time (s) & FEvals\\
\midrule
GA & 3 & 501 & 1.7 &  502 \\
GAB-B & 3 & \textbf{25} & \textbf{0.038} & \textbf{26.8} \\
GA & 7 & 72 & 0.46 &  81.4 \\
GAB-B & 7 & \textbf{26} & \textbf{0.21} & \textbf{34.6} \\
GA & 10 & 183 & 1.001 &  202.3 \\
GAB-B & 10 & \textbf{95} & \textbf{0.52} & \textbf{110.8} \\
GA & 13 & 291 & 1.55 &  317.6 \\
GAB-B & 13 & \textbf{135} & \textbf{1.198} & \textbf{159.6} \\
GA & 18 & N/A & N/A &  N/A \\
GAB-B & 18 & \textbf{681} & \textbf{9.37} & \textbf{711} \\
\bottomrule
\end{tabular}
\end{sc}
\end{small}
\end{center}
\vskip -0.1in
\end{table}

GA exhibited only semi-convergence at the 494th iteration for a problem size of 18 and never fully converged. Between iterations 220 and 817, GA intermittently reached the desired optimal fitness value of \$422; however, the highest average fitness value over 10 runs remained at \$421. The underlying reason is illustrated in Figure 6[i] and 6[j]: when GA achieved a fitness of \$422 within this iteration range, the standard deviation (Std Dev) of fitness was merely 1.6, as indicated by the nearly invisible Std Dev bands in the graph. This suggests that the search lacked sufficient exploratory force to drive it in the correct direction.
Conversely, GAB-B exhibited a turning point at iteration 51 but subsequently stagnated at suboptimal fitness values between \$416 and \$420 for nearly 500 iterations (from iteration 135 to 621) at problem size 18. Notably, GAB-B demonstrated a significantly larger Std Dev of fitness, ranging from 8.09 to 10.08 during this stagnation phase, as depicted by the prominent Std Dev band in Figure 6[j]. This further highlights the added flexibility and exploratory drive introduced by border trades in GA, enabling more effective navigation of a constrained and complex problem space.

\section{Conclusion}
This study introduced novel chromosome patterns in the breeding process of the Genetic Algorithm through the implementation of border trades. Empirical results indicate that incorporating border exchanges enhances convergence efficiency, particularly in large or complex problem spaces. While border trades introduce additional computational costs, this can be mitigated when the exploratory benefits outweigh the associated computational overhead. Future research could evaluate the comparative performance of GAB against alternative approaches, such as Genitor and Simulated Annealing, and exploring the potential benefits of improving GAB with the advantages of these approaches.

\newpage
\onecolumn
\section*{APPENDIX}

\subsection{Hyperparameter Tuning}
\subsubsection{Flip-Flop Hyperparameters}
For valid comparisons, each pair of GA and GAB results within the same problem space utilized identical hyperparameter candidates and thresholds for early stopping criteria, such as maximum attempts and iterations. In Table V, "Best Pop Size" denotes optimal population sizes; bolded fitness values in the last column highlight suboptimal solutions.

\begin{table}[h]
\caption{Optimal Hyperparameters for GA and GAB}
\vskip 0.15in
\begin{center}
\begin{small}
\begin{sc}
\begin{tabular}{lcccccr}
\toprule
 Algorithm & Size & Best Pop Size & Best Mutation Rate & Population Sizes & Mutation Rates & Best Fitness\\
\midrule
GA & 1,000 & 150 & 0.08 & [20, 50, 100, 150] & [0.1, 0.08] & \textbf{843}\\
GAB & 1,000 & 150 & 0.08 & [20, 50, 100, 150] & [0.1, 0.08] & 999\\
GA & 28 & 16 & 0.1 & [16, 18] & [0.1, 0.2] & \textbf{26}\\
GAB & 28 & 16 & 0.2 & [16, 18] & [0.1, 0.2] & 27\\
GA & 14 & 5 & 0.5 & [5, 10] & [0.4, 0.5] & 13\\
GAB & 14 & 5 & 0.5 & [5, 10] & [0.4, 0.5] & 13\\
GA & 7 & 3 & 0.4 & [3, 5] & [0.4, 0.5] & 6\\
GA & 7 & 3 & 0.5 & [3, 5] & [0.4, 0.5] & 6\\
GA & 7 & 5 & 0.4 & [3, 5] & [0.4, 0.5] & 6\\
GA & 7 & 5 & 0.5 & [3, 5] & [0.4, 0.5] & 6\\
GAB & 7 & 3 & 0.4 & [3, 5] & [0.4, 0.5] & 6\\
GAB & 7 & 3 & 0.5 & [3, 5] & [0.4, 0.5] & 6\\
GAB & 7 & 5 & 0.4 & [3, 5] & [0.4, 0.5] & 6\\
GAB & 7 & 5 & 0.5 & [3, 5] & [0.4, 0.5] & 6\\
\bottomrule
\end{tabular}
\end{sc}
\end{small}
\end{center}
\vskip -0.1in
\end{table}

The third and fourth columns, labeled "best population size" and "best mutation rate," represent the selected hyperparameters that yielded the highest fitness values with the lowest number of function evaluations (FEvals) and minimal wall-clock time in a single run (detailed code in Suplementary Materials). In cases where multiple hyperparameter combinations produced identical FEvals, iterations and run-time, all tied combinations are listed in the table. Notably, the GAB algorithm frequently achieved the target fitness value using multiple combinations of population sizes and mutation rates, whereas the GA algorithm typically corresponded to a single hyperparameter combination, offering limited flexibility. This observation highlights the increased adaptability introduced by border trades, which enhanced the algorithm's exploratory capabilities.

Due to the lack of convergent results of GA for problem sizes 28 and 1000 when using the original hyperparameter values for problem space 7, the population size range was expanded, and the mutation rate range was reduced. These adjustments led to an empirical increase in the highest fitness value from 25 to 26 in GA studies for problem size 28, although this still fell short of the theoretical maximum. Specifically, a population size of 16 and a mutation rate of 0.1 were the only combination that resulted in semi-convergence to a suboptimal fitness value of 26 for problem size 28. For problem size 1000, a population size of 150 and a mutation rate of 0.08 were the only settings that led to a suboptimal fitness value of 843. Section III.A provides an explanation for why GAB achieved convergence, whereas GA did not, even when both used the same parameters. Future research could investigate how further increasing population sizes and decreasing mutation rates might help mitigate premature convergence in larger problem spaces.

\newpage
\subsubsection{Job Scheduling Hyperparameters}
Due to the complexity of Job Scheduling with Breaks problems, the candidate population sizes were generally larger to achieve smaller convergence iterations. A comparison between small population sizes [2, 4, 5] and large population sizes [40, 50, 60] could be found in supplementary notebooks js-ga-13.ipynb and js-ga-retuned-13.ipynb. It is important to note that the best fitness values did not increase monotonically with problem size. This is attributable to the random generation of tasks at each problem size using a fixed seed, rather than incrementally appending tasks as the problem size increased. Similar to the previous appendix, in cases where multiple hyperparameter combinations produced identical FEvals, iterations and run-time, all tied combinations are listed in the table. See js-brute-force-verification.ipynb in supplementary materials for convergence verification. The max fitness that was less than the desired value is highlighted in the table.

\begin{table}[h]
\caption{Optimal Hyperparameters for GA and GAB-B}
\vskip 0.15in
\begin{center}
\begin{small}
\begin{sc}
\begin{tabular}{lcccccr}
\toprule
 Algorithm & Size & Best Pop Size & Best Mutation Rate & Population Sizes & Mutation Rates & Best Fitness\\
\midrule
GA & 18 & 60 & 0.08 & [40, 50, 60] & [0.07, 0.08] & \textbf{421}\\
GAB-B & 18 & 60 & 0.08 & [40, 50, 60] & [0.07, 0.08] & 422\\
GA & 13 & 60 & 0.08 & [40, 50, 60] & [0.07, 0.08] & 299\\
GAB-B & 13 & 50 & 0.08 &  [40, 50, 60] & [0.07, 0.08] & 299\\
GA & 10 & 60 & 0.08 & [40, 50, 60] & [0.07, 0.08] & 230\\
GAB-B & 10 & 40 & 0.07 & [40, 50, 60] & [0.07, 0.08] & 230\\
GA & 7 & 60 & 0.08 & [40, 50, 60] & [0.07, 0.08] & 161\\
GAB-B & 7 & 50 & 0.08 &  [40, 50, 60] & [0.07, 0.08] & 161\\
GA & 3 & 4 & 0.08 & [2, 4, 5] & [0.1, 0.08] & 180\\
GA & 3 & 4 & 0.1 & [2, 4, 5] & [0.1, 0.08] & 180\\
GAB-B & 3 & 5 & 0.08 & [2, 4, 5] & [0.1, 0.08] & 180\\
GAB-B & 3 & 5 & 0.1 & [2, 4, 5] & [0.1, 0.08] & 180\\
\bottomrule
\end{tabular}
\end{sc}
\end{small}
\end{center}
\vskip -0.1in
\end{table}

\newpage
\subsection{Additional Graphs of Job Scheduling Problems}

\subsubsection{Performance of Different Border Trading Strategies for Problem Size = 108} 
The fitness, FEvals and wall-clock time comparison among various border trading strategies.

\begin{figure}[ht] 
    \begin{subfigure}{0.15\textwidth}
       \includegraphics[width=\linewidth]{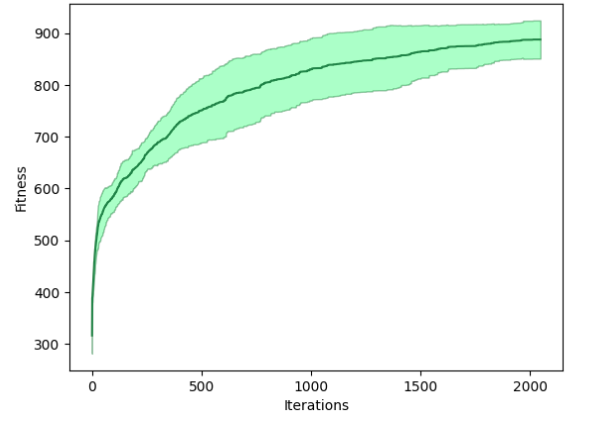}
       \caption{Fitness by Iteration of GA}       
   \end{subfigure}
\hfill  
   \begin{subfigure}{0.15\textwidth}
       \includegraphics[width=\linewidth]{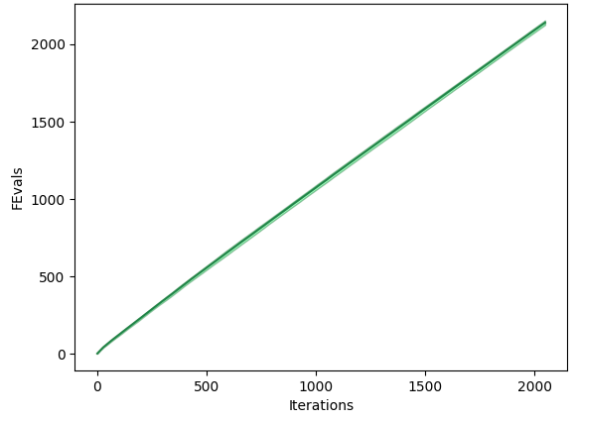}
       \caption{FEvals by Iteration of GA}
   \end{subfigure}
\hfill 
   \begin{subfigure}{0.15\textwidth}
       \includegraphics[width=\linewidth]{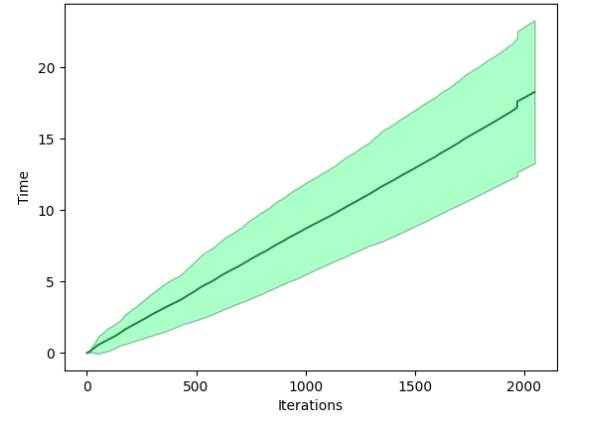}
       \caption{Time by Iteration of GA}
   \end{subfigure}
\hfill 
   \begin{subfigure}{0.15\textwidth}
       \includegraphics[width=\linewidth]{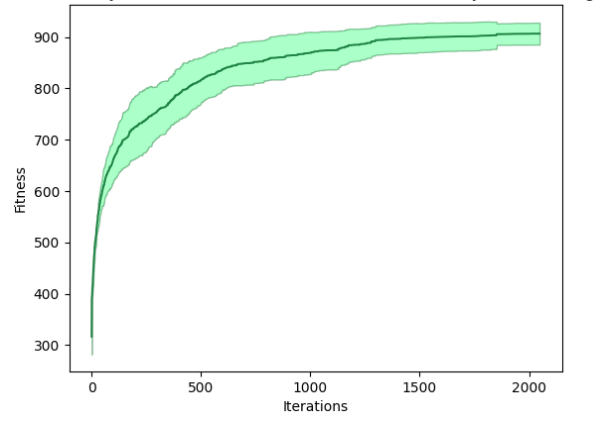}
       \caption{Fitness by Iteration of GAB-A1}       
   \end{subfigure}
\hfill  
   \begin{subfigure}{0.15\textwidth}
       \includegraphics[width=\linewidth]{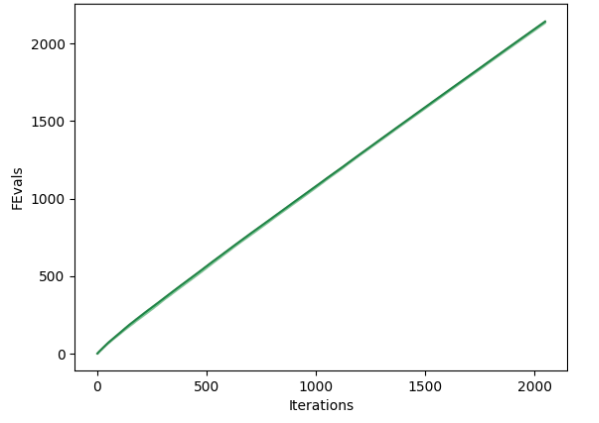}
       \caption{FEvals by Iteration of GAB-A1}
   \end{subfigure}
\hfill 
   \begin{subfigure}{0.15\textwidth}
       \includegraphics[width=\linewidth]{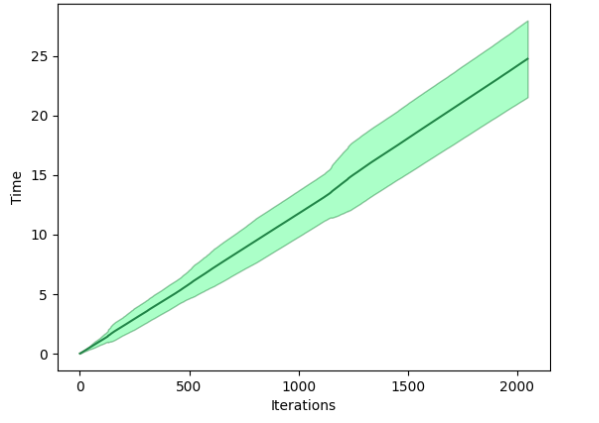}
       \caption{Time by Iteration of GAB-A1}
   \end{subfigure}
\hfill 
   \begin{subfigure}{0.15\textwidth}
       \includegraphics[width=\linewidth]{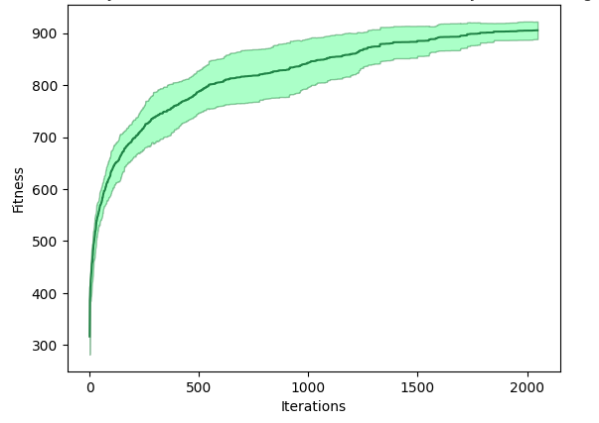}
       \caption{Fitness by Iteration of GAB-A2}       
   \end{subfigure}
\hfill  
   \begin{subfigure}{0.15\textwidth}
       \includegraphics[width=\linewidth]{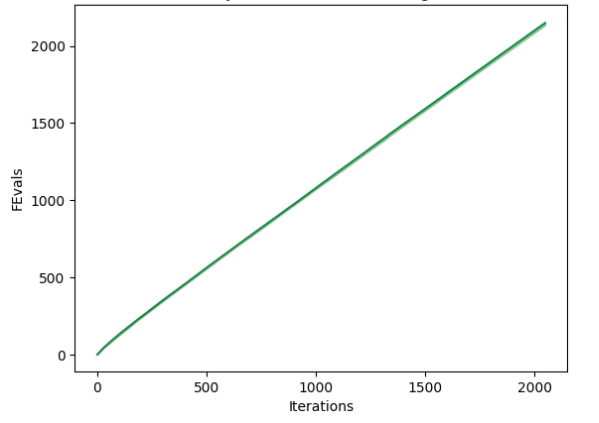}
       \caption{FEvals by Iteration of GAB-A2}
   \end{subfigure}
\hfill 
   \begin{subfigure}{0.15\textwidth}
       \includegraphics[width=\linewidth]{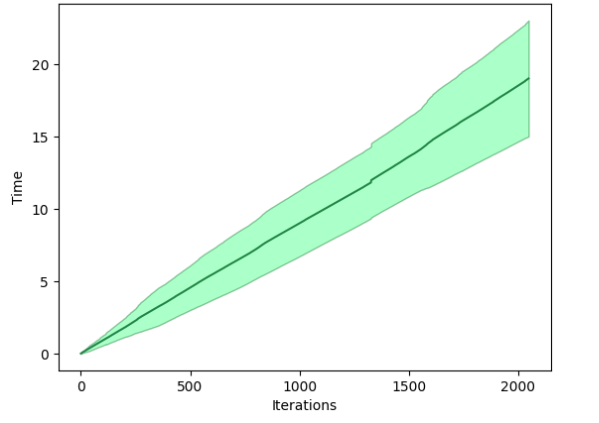}
       \caption{Time by Iteration of GAB-A2}
   \end{subfigure}
\hfill
   \begin{subfigure}{0.15\textwidth}       
       \includegraphics[width=\linewidth]{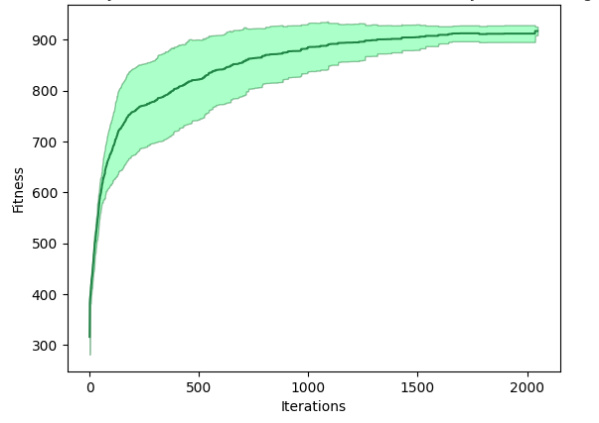}
       \caption{Fitness by Iteration of GAB-B}
   \end{subfigure}
\hfill  
   \begin{subfigure}{0.15\textwidth}
       \includegraphics[width=\linewidth]{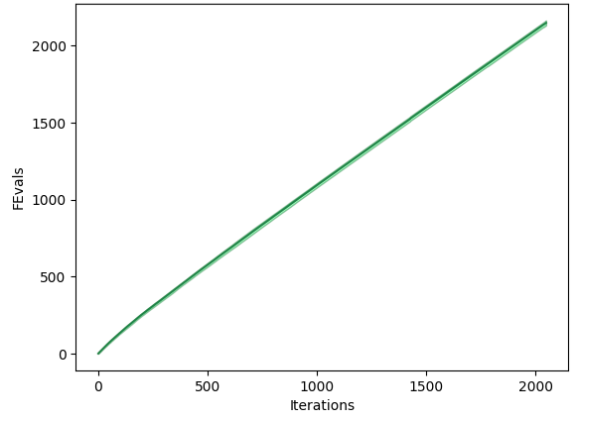}
       \caption{FEvals by Iteration of GAB-B}
   \end{subfigure}
\hfill 
   \begin{subfigure}{0.15\textwidth}
       \includegraphics[width=\linewidth]{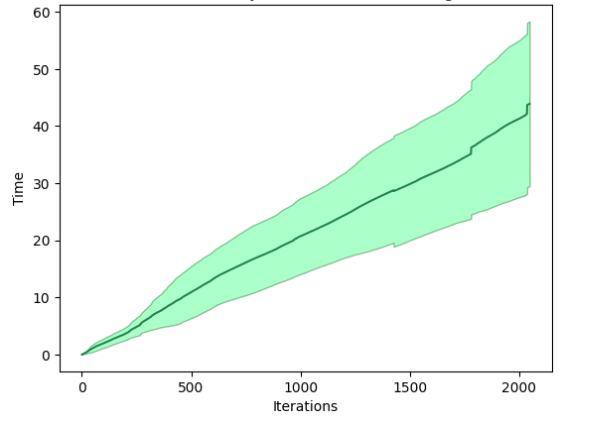}
       \caption{Time by Iteration of GAB-B}
   \end{subfigure} 
\hfill
   \begin{subfigure}{0.15\textwidth}
       \includegraphics[width=\linewidth]{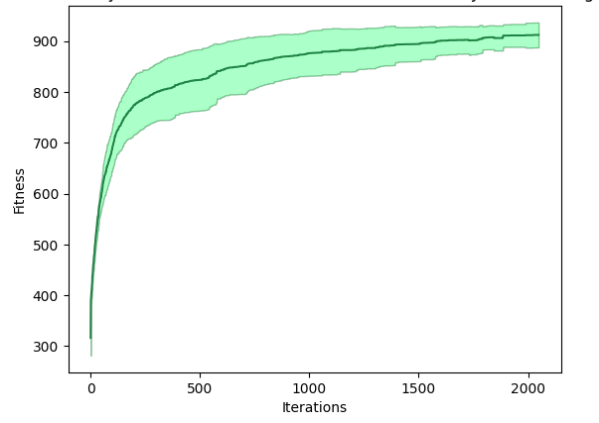}
       \caption{Fitness by Iteration of GAB-C1}       
   \end{subfigure}
\hfill  
   \begin{subfigure}{0.15\textwidth}
       \includegraphics[width=\linewidth]{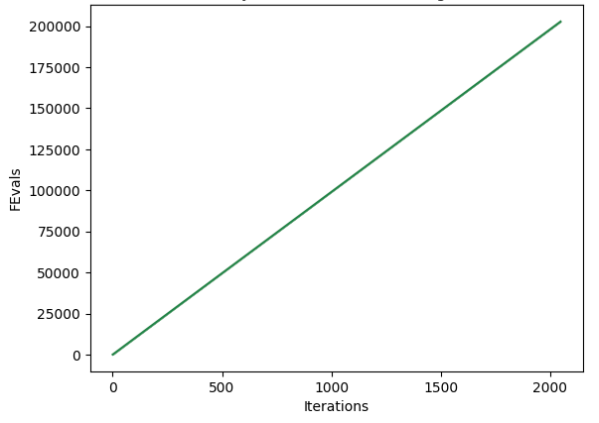}
       \caption{FEvals by Iteration of GAB-C1}
   \end{subfigure}
\hfill 
   \begin{subfigure}{0.15\textwidth}
       \includegraphics[width=\linewidth]{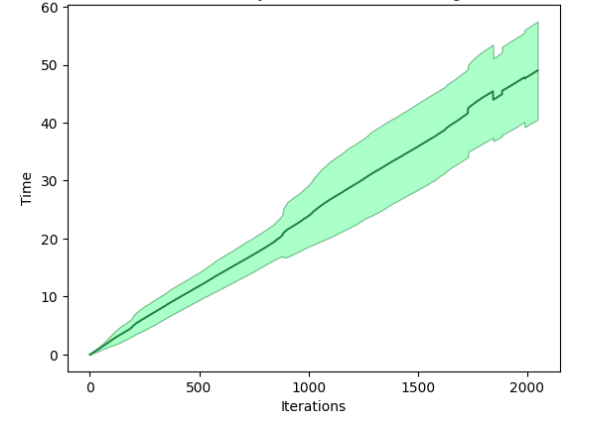}
       \caption{Time by Iteration of GAB-C1}
   \end{subfigure}
\hfill 
   \begin{subfigure}{0.15\textwidth}       
       \includegraphics[width=\linewidth]{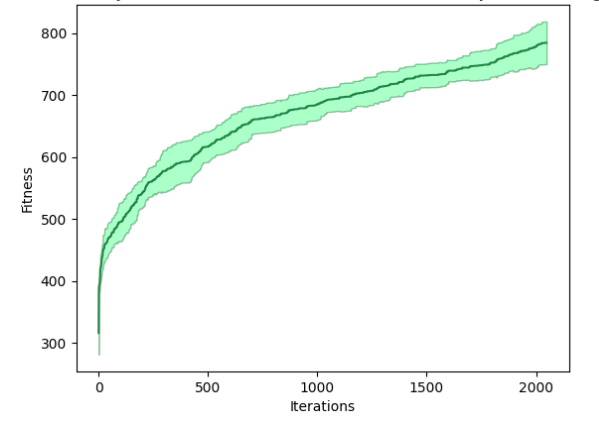}
       \caption{Fitness by Iteration of GAB-C2}
   \end{subfigure}
\hfill  
   \begin{subfigure}{0.15\textwidth}
       \includegraphics[width=\linewidth]{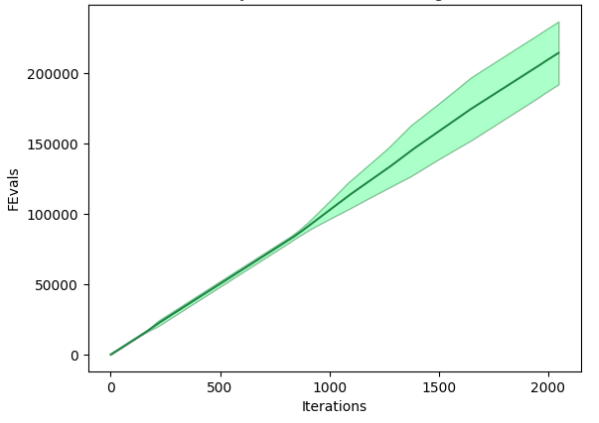}
       \caption{FEvals by Iteration of GAB-C2}
   \end{subfigure}
\hfill 
   \begin{subfigure}{0.15\textwidth}
       \includegraphics[width=\linewidth]{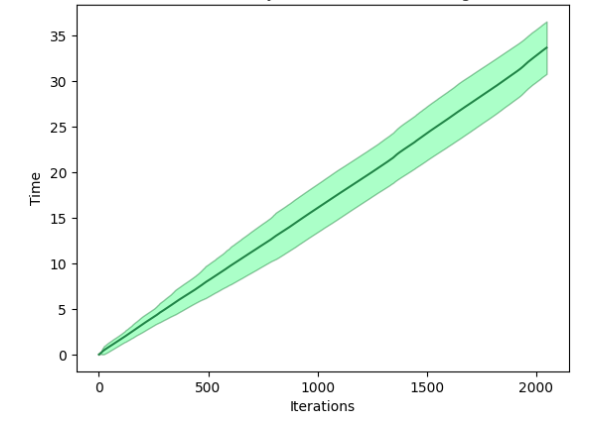}
       \caption{Time by Iteration of GAB-C2}
   \end{subfigure} 
\caption{Performance of Different Border Trading Strategies for Problem Size = 108}   
\end{figure}

\newpage
\subsubsection{Performance Comparisons Between GA and GAB-B in Job Scheduling with Breaks Problems}
At various problem sizes, compare the performance of GA and GAB-B.

\begin{figure}[ht]
\vskip 0.2in
\centering
\begin{subfigure}[b]{0.15\textwidth}
         \centering
         \includegraphics[width=\linewidth]{js-ga-3-fitness.png}
         \caption{GA Fitness, Size=3}         
     \end{subfigure}
     \hfill
     \begin{subfigure}[b]{0.15\textwidth}
         \centering
         \includegraphics[width=\linewidth]{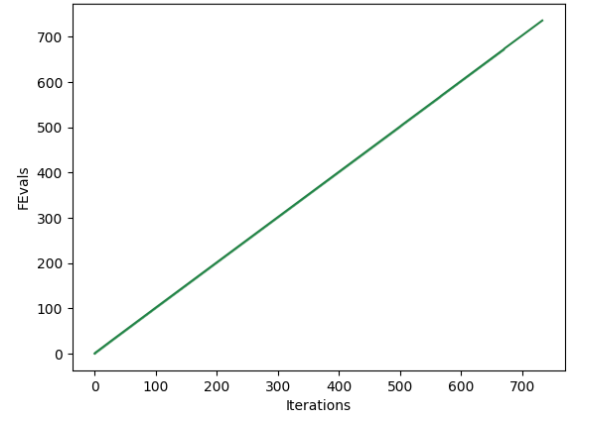}
         \caption{GA FEvals, Size=3}         
     \end{subfigure}
     \hfill
     \begin{subfigure}[b]{0.15\textwidth}
         \centering
         \includegraphics[width=\linewidth]{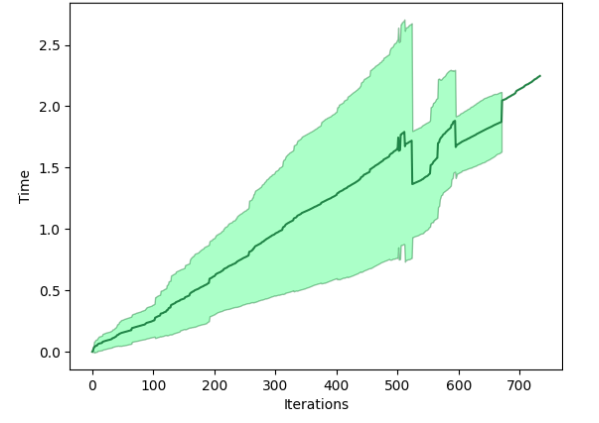}
         \caption{GA Wall-Clock Time, Size=3}         
     \end{subfigure}
     \hfill
     \begin{subfigure}[b]{0.15\textwidth}
         \centering
         \includegraphics[width=\linewidth]{js-gab-3-fitness.png}
         \caption{GAB Fitness, Size=3}
     \end{subfigure}
     \hfill
     \begin{subfigure}[b]{0.15\textwidth}
         \centering
         \includegraphics[width=\linewidth]{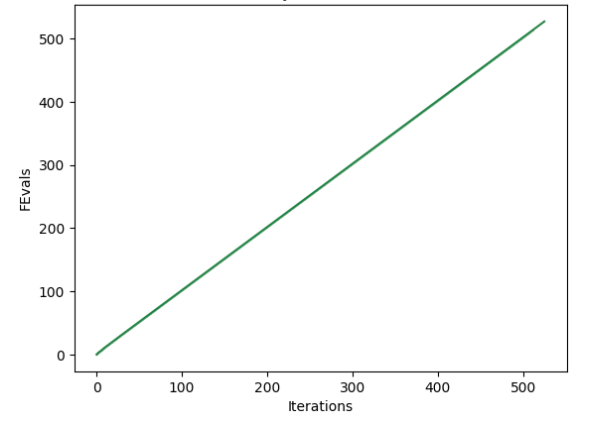}
         \caption{GAB FEvals, Size=3}
     \end{subfigure}
     \hfill
     \begin{subfigure}[b]{0.15\textwidth}
         \centering
         \includegraphics[width=\linewidth]{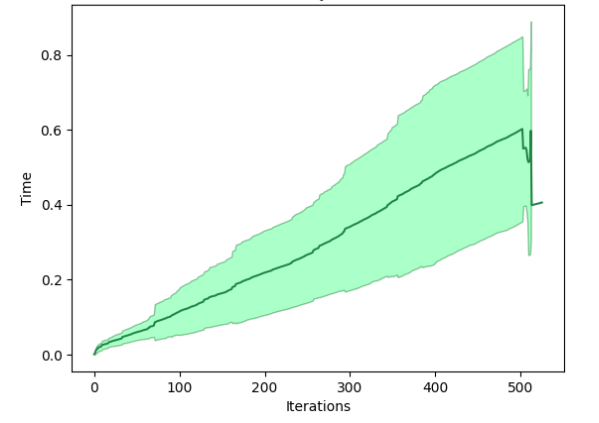}
         \caption{GAB Time, Size=3}
     \end{subfigure}
     \hfill
     \begin{subfigure}[b]{0.15\textwidth}
         \centering
         \includegraphics[width=\linewidth]{js-ga-7-fitness.png}
         \caption{GA Fitness, Size=7}         
     \end{subfigure}
     \hfill
     \begin{subfigure}[b]{0.15\textwidth}
         \centering
         \includegraphics[width=\linewidth]{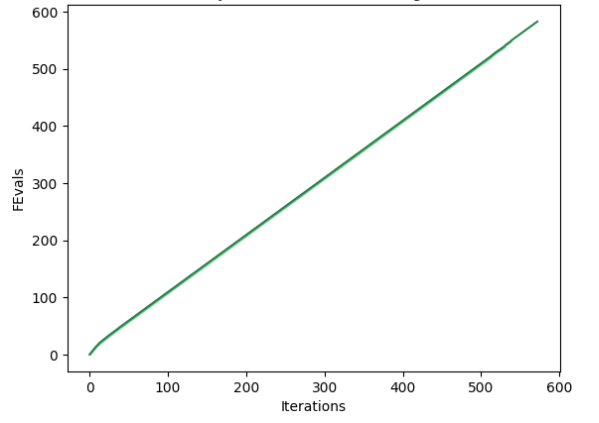}
         \caption{GA FEvals, Size=7}
     \end{subfigure}
     \hfill
     \begin{subfigure}[b]{0.15\textwidth}
         \centering
         \includegraphics[width=\linewidth]{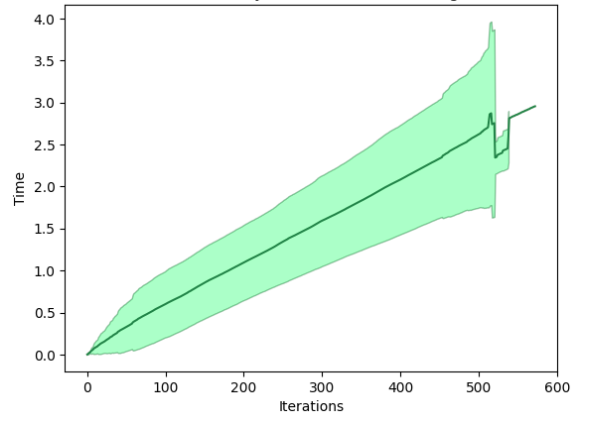}
         \caption{GA Wall-Clock Time, Size=7}
     \end{subfigure}
     \hfill
     \begin{subfigure}[b]{0.15\textwidth}
         \centering
         \includegraphics[width=\linewidth]{js-gab-7-fitness.png}
         \caption{GAB Fitness, Size=7}
     \end{subfigure}
     \hfill
     \begin{subfigure}[b]{0.15\textwidth}
         \centering
         \includegraphics[width=\linewidth]{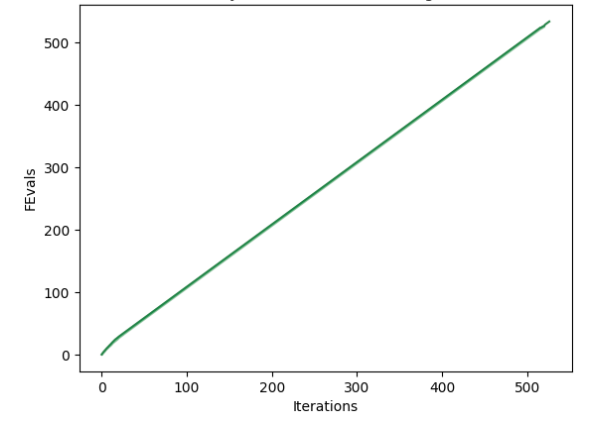}
         \caption{GAB FEvals, Size=7}
     \end{subfigure}
     \hfill
     \begin{subfigure}[b]{0.15\textwidth}
         \centering
         \includegraphics[width=\linewidth]{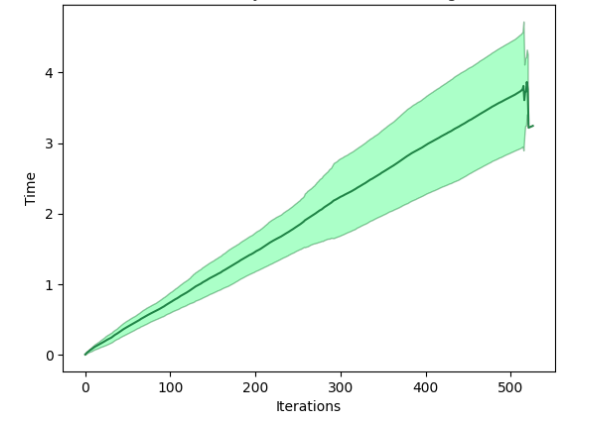}
         \caption{GAB Time, Size=7}
     \end{subfigure}
     \hfill
     \begin{subfigure}[b]{0.15\textwidth}
         \centering
         \includegraphics[width=\linewidth]{js-ga-10-fitness.png}
         \caption{GA Fitness, Size=10}         
     \end{subfigure}
     \hfill
     \begin{subfigure}[b]{0.15\textwidth}
         \centering
         \includegraphics[width=\linewidth]{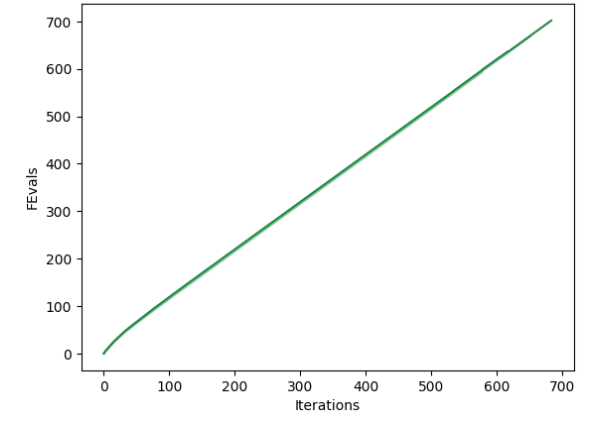}
         \caption{GA FEvals, Size=10}
     \end{subfigure}
     \hfill
     \begin{subfigure}[b]{0.15\textwidth}
         \centering
         \includegraphics[width=\linewidth]{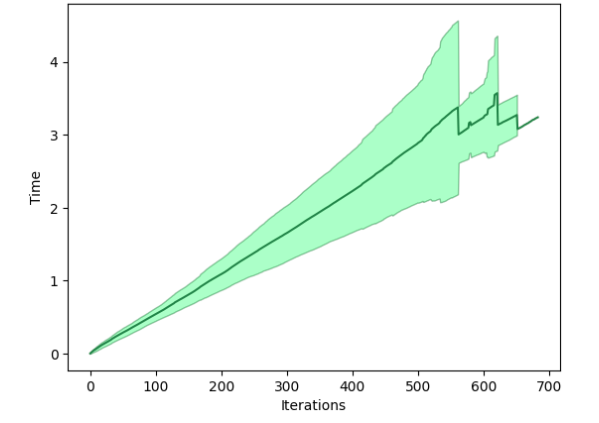}
         \caption{GA Wall-Clock Time, Size=10}
     \end{subfigure}
     \hfill
     \begin{subfigure}[b]{0.15\textwidth}
         \centering
         \includegraphics[width=\linewidth]{js-gab-10-fitness.png}
         \caption{GAB Fitness, Size=10}
     \end{subfigure}
     \hfill
     \begin{subfigure}[b]{0.15\textwidth}
         \centering
         \includegraphics[width=\linewidth]{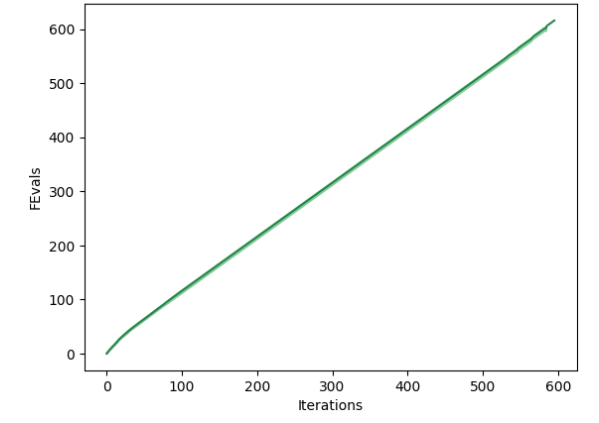}
         \caption{GAB FEvals, Size=10}
     \end{subfigure}
     \hfill
     \begin{subfigure}[b]{0.15\textwidth}
         \centering
         \includegraphics[width=\linewidth]{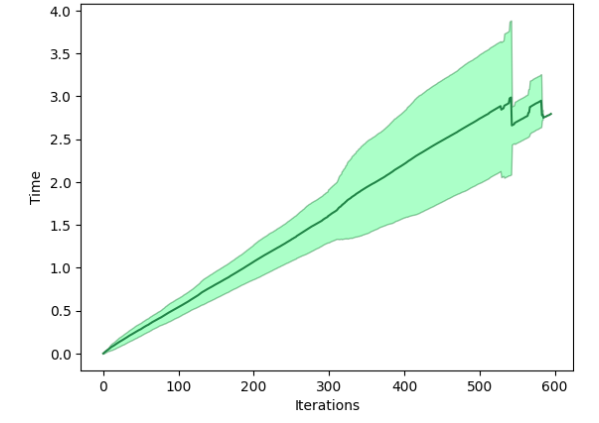}
         \caption{GAB Time, Size=10}
     \end{subfigure}
     \hfill
          \begin{subfigure}[b]{0.15\textwidth}
         \centering
         \includegraphics[width=\linewidth]{js-ga-13-fitness.png}
         \caption{GA Fitness, Size=13}         
     \end{subfigure}
     \hfill
     \begin{subfigure}[b]{0.15\textwidth}
         \centering
         \includegraphics[width=\linewidth]{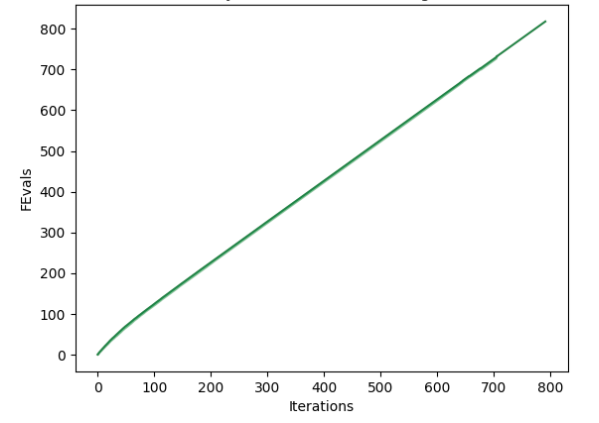}
         \caption{GA FEvals, Size=13}
     \end{subfigure}
     \hfill
     \begin{subfigure}[b]{0.15\textwidth}
         \centering
         \includegraphics[width=\linewidth]{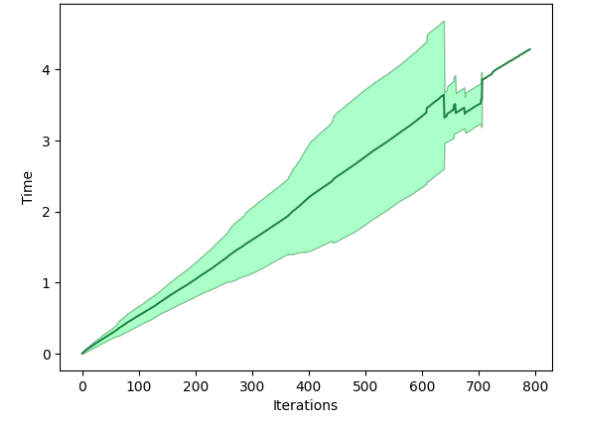}
         \caption{GA Wall-Clock Time, Size=13}
     \end{subfigure}
     \hfill
     \begin{subfigure}[b]{0.15\textwidth}
         \centering
         \includegraphics[width=\linewidth]{js-gab-13-fitness.png}
         \caption{GAB Fitness, Size=13}
     \end{subfigure}
     \hfill
     \begin{subfigure}[b]{0.15\textwidth}
         \centering
         \includegraphics[width=\linewidth]{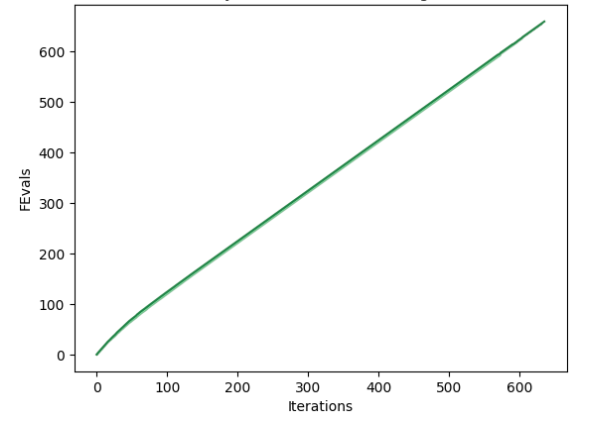}
         \caption{GAB FEvals, Size=13}
     \end{subfigure}
     \hfill
     \begin{subfigure}[b]{0.15\textwidth}
         \centering
         \includegraphics[width=\linewidth]{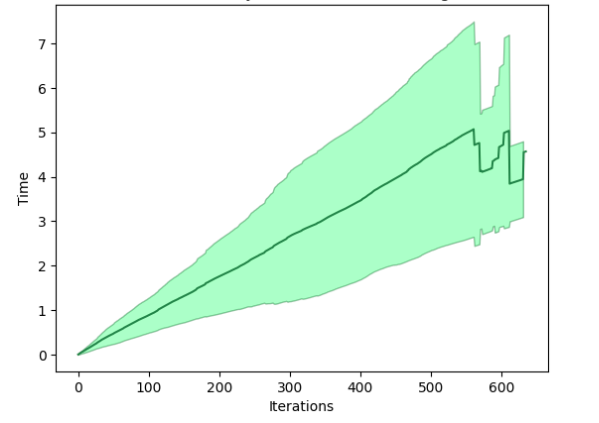}
         \caption{GAB Time, Size=13}
     \end{subfigure}
     \hfill
     \begin{subfigure}[b]{0.15\textwidth}
         \centering
         \includegraphics[width=\linewidth]{js-ga-18-fitness.png}
         \caption{GA Fitness, Size=18}         
     \end{subfigure}
     \hfill
     \begin{subfigure}[b]{0.15\textwidth}
         \centering
         \includegraphics[width=\linewidth]{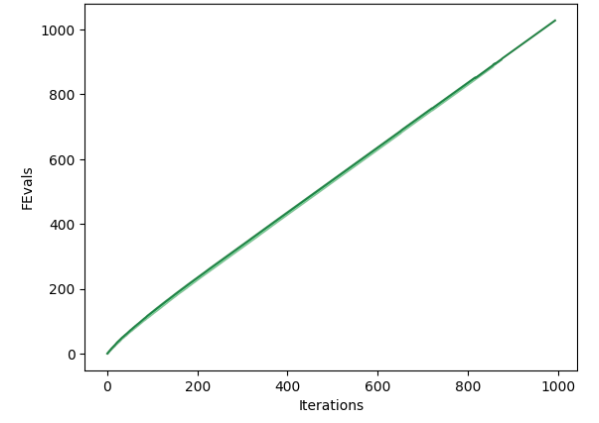}
         \caption{GA FEvals, Size=18}
     \end{subfigure}
     \hfill
     \begin{subfigure}[b]{0.15\textwidth}
         \centering
         \includegraphics[width=\linewidth]{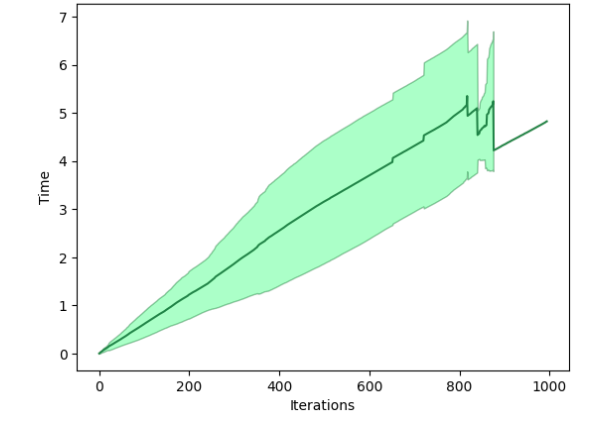}
         \caption{GA Wall-Clock Time, Size=18}
     \end{subfigure}
     \hfill
     \begin{subfigure}[b]{0.15\textwidth}
         \centering
         \includegraphics[width=\linewidth]{js-gab-18-fitness.png}
         \caption{GAB Fitness, Size=18}
     \end{subfigure}
     \hfill
     \begin{subfigure}[b]{0.15\textwidth}
         \centering
         \includegraphics[width=\linewidth]{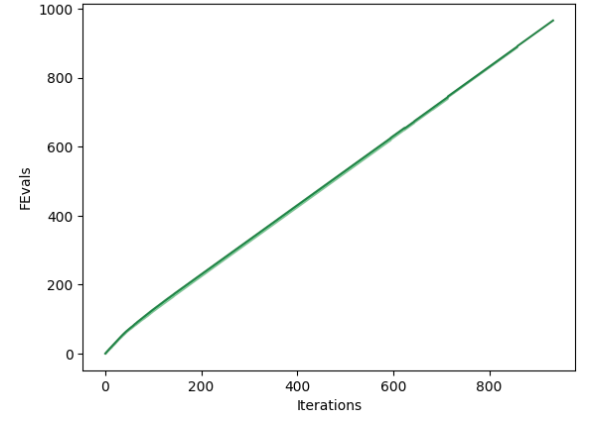}
         \caption{GAB FEvals, Size=18}
     \end{subfigure}
     \hfill
     \begin{subfigure}[b]{0.15\textwidth}
         \centering
         \includegraphics[width=\linewidth]{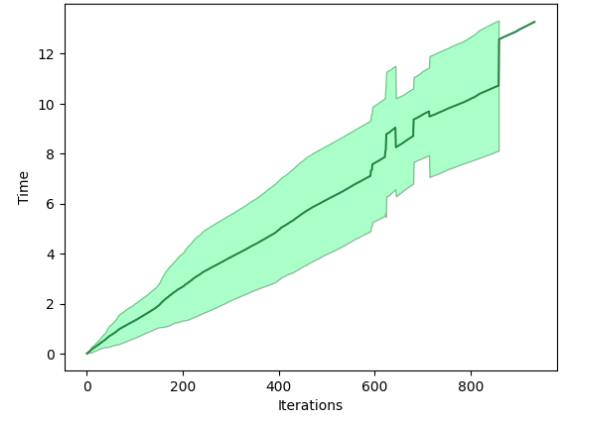}
         \caption{GAB Time, Size=18}
     \end{subfigure}
     \hfill
     \caption{Performance Comparisons Between GA and GAB-B in Job Scheduling with Breaks Problems. Left: curves for GA; Right: curves for GAB-B.}   
\vskip -0.2in
\end{figure}

\newpage
\subsection{Simulated Annealing Performance in Flip-Flop Optimization}
\begin{figure}[ht]
\vskip 0.2in
\centering
\begin{subfigure}[b]{0.24\textwidth}
         \centering
         \includegraphics[width=\linewidth]{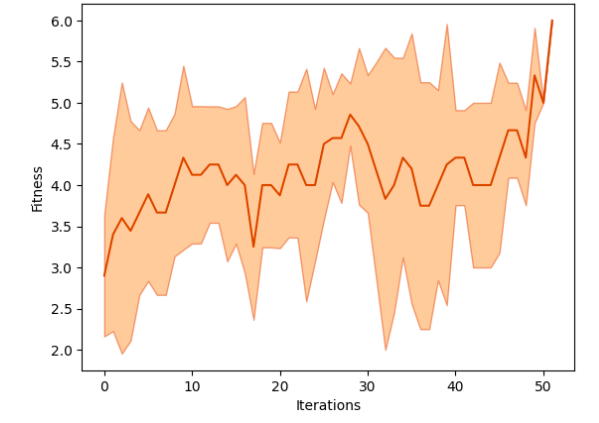}
         \caption{Problem Size=7}         
     \end{subfigure}
     \hfill
     \begin{subfigure}[b]{0.24\textwidth}
         \centering
         \includegraphics[width=\linewidth]{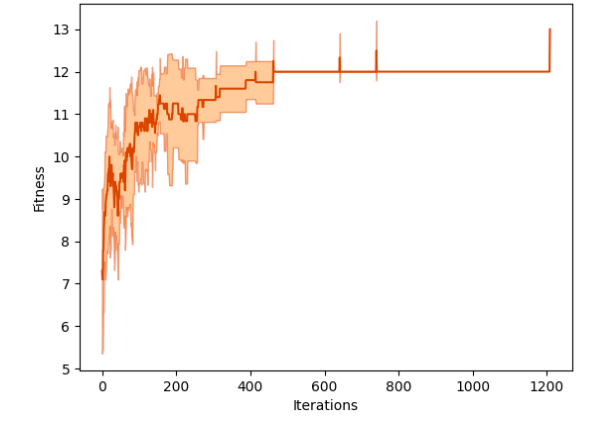}
         \caption{Problem Size=14}         
     \end{subfigure}
     \hfill
     \begin{subfigure}[b]{0.24\textwidth}
         \centering
         \includegraphics[width=\linewidth]{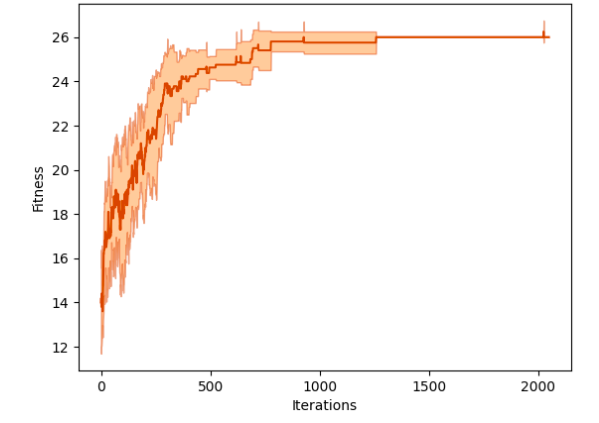}
         \caption{Problem Size=28}         
     \end{subfigure}
     \hfill
     \begin{subfigure}[b]{0.24\textwidth}
         \centering
         \includegraphics[width=\linewidth]{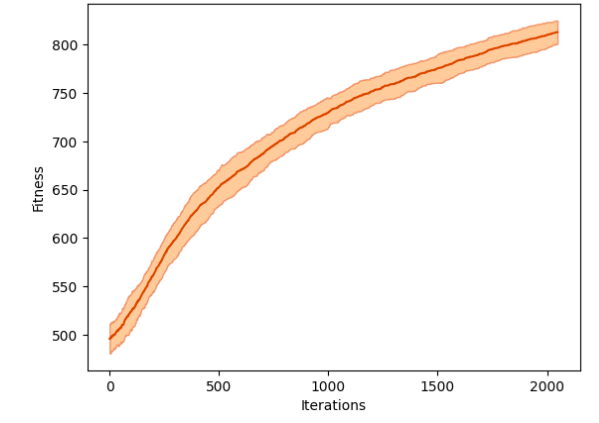}
         \caption{Problem Size=1,000}
     \end{subfigure}
     \hfill
     \caption{Performance of Simulated Annealing (SA) in Flip Flop Problems.}   
\vskip -0.2in
\end{figure}

\end{document}